%% file: main.tex
\documentclass{article}
\usepackage{neurips_2024}

\input{command/math_commands}

\usepackage[utf8]{inputenc}
\usepackage[T1]{fontenc}
\usepackage{microtype}
\usepackage{amsmath}
\usepackage{xspace}
\usepackage{booktabs}
\usepackage{bookmark}
\usepackage{mathtools}
\usepackage{nccmath}
\usepackage{setspace}
\usepackage{wrapfig, lipsum}
\usepackage{soul}
\usepackage{dsfont}
\usepackage{enumerate}
\usepackage{enumitem}
\usepackage{amsfonts}
\usepackage{bbm}
\usepackage[Symbol]{upgreek}
\usepackage{lscape}
\usepackage{caption}
\usepackage{balance}
\usepackage{float}
\usepackage{algorithm}
\usepackage{algpseudocode}
\usepackage{wasysym}
\usepackage[table, xcdraw, dvipsnames]{xcolor}
\usepackage{multirow}
\usepackage{array, boldline, rotating}
\usepackage{amssymb}
\usepackage{pifont}
\usepackage{framed}
\usepackage{hyperref}
\usepackage{url}
\usepackage{nicefrac}
\usepackage{graphicx}
\usepackage{adjustbox}
\usepackage{blindtext}

\usepackage{amsthm}
\usepackage{subcaption}

\makeatletter
\let\newfloat\newfloat@ltx
\makeatother

\hypersetup{
  colorlinks   = true, %
  urlcolor     = ForestGreen, %
  linkcolor    = ForestGreen, %
  citecolor   = cyan %
}

\theoremstyle{plain}
\theoremstyle{definition}

\usepackage[capitalize,noabbrev]{cleveref}
\crefname{section}{Sec.}{Secs.}
\Crefname{section}{Section}{Sections}
\Crefname{table}{Table}{Tables}
\crefname{table}{Tab.}{Tabs.}

\def\equationautorefname~#1\null{Eq.~(#1)\null}

\newcommand{\alg}{\code{VACoDe}\xspace}

\newcommand{\mytitle}{VACoDe: Visual Augmented Contrastive Decoding}

\newcommand{\mc}[1]{\mathcal{#1}}

\renewcommand*\eqref[1]{(\ref{#1})}

\newcommand{\eg}{\emph{e.g.,~}}
\newcommand{\ie}{\emph{i.e.,~}}
\newcommand{\algcomment}[1]{\hfill\textcolor{SkyBlue}{$\triangleright$ #1}}
\newcommand{\bignorm}[1]{\left\lVert#1\right\rVert}
\newcommand{\myparagraph}[1]{\vspace{0.07cm}\noindent\textbf{#1}~}
\def\code#1{\texttt{#1}}

\newcommand{\thickhline}{\hlineB{4}}

\definecolor{LightCyan}{rgb}{0.88,1,1}
\definecolor{LightGray}{gray}{0.9}
\definecolor{LightRed}{rgb}{1,0.95,0.95}
\definecolor{Red}{rgb}{0.95, 0.55, 0.6}
\definecolor{Skyblue}{rgb}{0.6, 0.6, 0.95 }
\definecolor{SKY}{rgb}{0.9, 0.99, 1.0 }
\definecolor{shadecolor}{named}{LightGray}

\NewDocumentCommand{\supptitle}{s}{
\onecolumn
\begin{center}
    \rule{\textwidth}{0.03cm}\\[0.1cm]
    -Supplementary Material-\\[0.2cm]
    {\Large 
        \textbf{\mytitle }
    }\\
    \rule{\textwidth}{0.03cm}\\[0.2cm]
\end{center}
}

\title{\mytitle}

\author{%
  Sihyeon Kim\thanks{Equal Contribution}, \quad Boryeong Cho$^*$, \quad Sangmin Bae \\
  KAIST AI\\
  \texttt{\{sihk, venntum, bsmn0223\}@kaist.ac.kr} \\
  \AND
  Sumyeong Ahn\thanks{Corresponding Author} \\
  CSE, Michigan State University \\
  \texttt{sumyeong@msu.edu} \\
  \And
  Se-Young Yun$^\dag$\\
  KAIST AI \\
  \texttt{yunseyoung@kaist.ac.kr} \\
}

\begin{document}

\maketitle

\input{main/00_abstract}

\input{main/01_intro}

\input{main/02_prelim}
\input{main/03_method}

\input{main/04_exp}

\input{main/05_related}

\input{main/06_conclusion}

\bibliography{main}
\bibliographystyle{plain}

\clearpage
\appendix{\input{main/appendix}}

\end{document}

%% file: command/math_commands.tex
\usepackage{amsmath,amsfonts,bm}

\def\eqref#1{equation~\ref{#1}}

\def\1{\bm{1}}

\DeclareMathAlphabet{\mathsfit}{\encodingdefault}{\sfdefault}{m}{sl}
\SetMathAlphabet{\mathsfit}{bold}{\encodingdefault}{\sfdefault}{bx}{n}

\DeclareMathOperator*{\argmax}{arg\,max}

%% file: main/00_abstract.tex
\begin{abstract}

Despite the astonishing performance of recent Large Vision-Language Models (LVLMs), these models often generate inaccurate responses. To address this issue, previous studies have focused on mitigating hallucinations by employing contrastive decoding (CD) with augmented images, which amplifies the contrast with the original image. However, these methods have limitations, including reliance on a single augmentation, which is restrictive for certain tasks, as well as the high cost of using external knowledge. In this study, we address these limitations by exploring how to utilize multiple image augmentations. Through extensive experiments, we observed that different augmentations produce varying levels of contrast depending on the task. Based on this observation, we introduce a novel method called \alg, Visual Augmented Contrastive Decoding. This method adaptively selects the augmentation with the highest contrast for each task using the proposed softmax distance metric. Our empirical tests show that \alg outperforms previous methods and improves output quality in various vision-language tasks. Additionally, \alg can be universally applied across different model types and sizes without additional training or the use of external models and data.

\end{abstract}

%% file: main/01_intro.tex
\section{Introduction}
\label{sec:intro}

Pre-trained Large Vision-Language Models (LVLMs)~\cite{liu2024visual, ye2023mplug, zhu2023minigpt, dai2024instructblip, li2022blip, li2023blip, radford2021learning} have gained prominence due to their capability to understand multiple data formats, especially vision and language, simultaneously. These models have demonstrated exceptional performance in various tasks like zero-shot image classification~\cite{radford2021learning, yao2021filip}, image-text retrieval~\cite{yao2021filip, li2022blip}, visual question answering~\cite{dai2024instructblip, liu2024visual}, and image captioning~\cite{li2022blip, li2023blip}. Different from earlier encoder-based models like CLIP~\cite{radford2021learning}, most recent large-scale VLMs, such as LLaVA~\cite{liu2024visual}, MPlugOWL~\cite{ye2023mplug}, MiniGPT-4~\cite{zhu2023minigpt}, and InstructBLIP~\cite{dai2024instructblip}, utilize autoregressive transformers to expand their functionality, enabling them to generate more complex outputs. %

However, language decoders sometimes produce incorrect outputs, a phenomenon often called hallucination. Among various methodologies~\cite{wei2022chain, rose2023visual, shao2024visual}, one promising approach is contrastive decoding (CD)~\cite{li-etal-2023-contrastive}, which generates final answers by examining candidate responses and leveraging their contrastiveness.  In detail, they operate in two stages: (1) generating output distributions given original and contrastive prompts each, and (2) subtracting the two output distributions to reduce the likelihood of hallucinated tokens. The effectiveness of this approach depends on how well the contrasting prompts are constructed. While it is relatively straightforward to create contrastive text prompts by replacing original words with their opposites or random words~\cite{instructivedecoding}, image prompts require a more deliberate approach, as there is no clearly defined augmentation strategy that creates contrastive information.

\begin{figure}[t]
    \centering
    \vspace{-5pt}
    \includegraphics[width=0.83\textwidth]{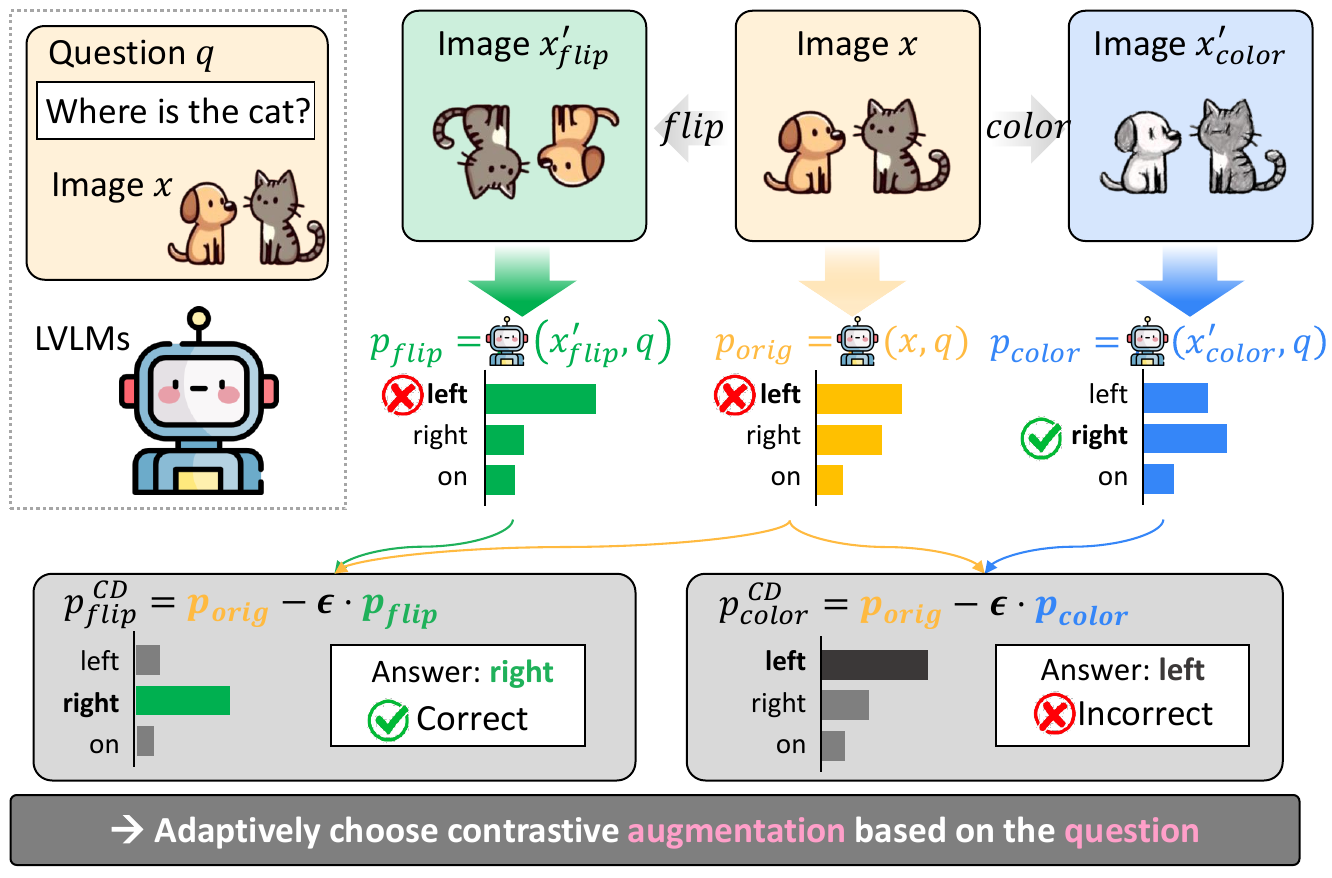}
    \caption{Overview of the problem we focus on: 
    When dealing with LVLMs, selecting the appropriate augmentation for each query is crucial to enhance decoding performance. 
    For example, if the question is ``Where is the cat?'' and the correct answer is \textit{right}, applying flip augmentation can alter the input image, resulting in a contrastive answer, \textit{left}. This contrastive information is beneficial for increasing the answer's probability when using CD. Conversely, using color augmentation for this question is unsuitable, as it does not generate contrastive output distributions. Therefore, the main challenge is how to adaptively select the most effective augmentation to improve CD performance in LVLMs.}
    \label{fig:intro}
    \vspace{-15pt}
\end{figure}

There have been a few works on generating contrastive images~\cite{leng2023mitigating, wan2024contrastive, chen2024halc}, which aim to increase sample variance by manipulating features in images through the addition of noise or cropping. In some cases, it has been shown that generating contrastive images using their static method can effectively modify features to create contrastiveness. 
However, applying fixed augmentations to all samples cannot always guarantee contrastive images, as the salient features in an image may vary depending on the text question in vision-language tasks.

We illustrate an example in~\autoref{fig:intro}, where the model receives position-related question ``\emph{where is the cat?},'' and generates the \emph{incorrect} output, \textit{left}. If a position-related augmentation such as flipping is applied to a given image, the output distribution would likely be heavily skewed towards \emph{left}. Thus, the model can generate the correct answer with the CD method by decreasing the probability of the hallucinated token. However, when applying color augmentation, where color is a less relevant feature to the target question, the color-augmented output distribution may be similar to the output distribution from the original image. 
Subsequently, the contrastive decoded logit with color augmentation may still have the wrong answer. 
In light of this, to generate appropriate and sufficient contrastiveness to ensure the model provides the correct answer, selecting the proper augmentation operation is significantly required.

\myparagraph{Contributions.} In this paper, we address the challenge of enhancing CD performance by formulating the selection of the proper augmentation. Our contributions are summarized as follows:
\begin{itemize}[leftmargin=15pt]
    \item We explore the effect of visual augmentation on LVLMs. Our findings indicate that each augmentation has a distinct impact, altering the output distribution of VLMs and subsequently affecting the response. From the CD perspective, the choice of proper augmentation is critical: selecting contrastive augmentations that introduce beneficial contrast can enhance performance, while persistent augmentations can lead to a decline in performance.
    \item Based on the findings, we introduce an algorithm called \alg that selects the most contrastive augmentation to empower CD capability without additional training or using external models. The algorithm consists of three main steps: (1) provide various types of augmented images to VLMs and generate multiple outputs. (2) Assess the difference between the original output distribution and the augmented output distributions. (3) Identify the most contrastive output, characterized by the largest distance gaps, and produce the final output by CD.
    \item Extensive empirical results verify that the proposed decoding method is superior to previous decoding techniques in VLMs.
    Furthermore, we observe evidence of why those augmentations work effectively in the contrastive decoding mechanism.
\end{itemize}

%% file: main/02_prelim.tex
\section{Preliminaries}
\label{sec:prelim}

Here, we provide a concise summary of background information to aid in understanding this research. We specifically discuss Vision-Language Models, visual data augmentation, and contrastive decoding.

\myparagraph{Generative Large Vision-Language Models (LVLMs).}
LVLMs are among the most prominent multi-modality models. They process pairs of input image $v$ and text (\eg question) $q$, denoted as $(v,q)$, and generate answers by utilizing the visual information within $v$. 
In this paper, we primarily focus on generative LVLMs, rather than CLIP-like~\cite{radford2021learning} models. These generative LVLMs produce tokens one at a time in sequence similar to LLMs. %
The mathematical expression for this process is:
\begin{equation*}
    y_{t} \sim p(y_{t} | v, q, y_{<t}).
\end{equation*}
Here, $p(\cdot)$ represents the softmax of the output of the vocabulary set, and $y_{<t}$ denotes the tokens generated up to but not including the timestamp $t$. Like LLMs, LVLMs are also prone to hallucination~\citep{li2023evaluating,liu2023mitigating,tong2024eyes}, where the model erroneously assigns higher probabilities to tokens that do not factually exist in the provided image.

\begin{figure}[h]
    \begin{subfigure}[b]{0.11\textwidth}
        \includegraphics[width=1.0\textwidth]{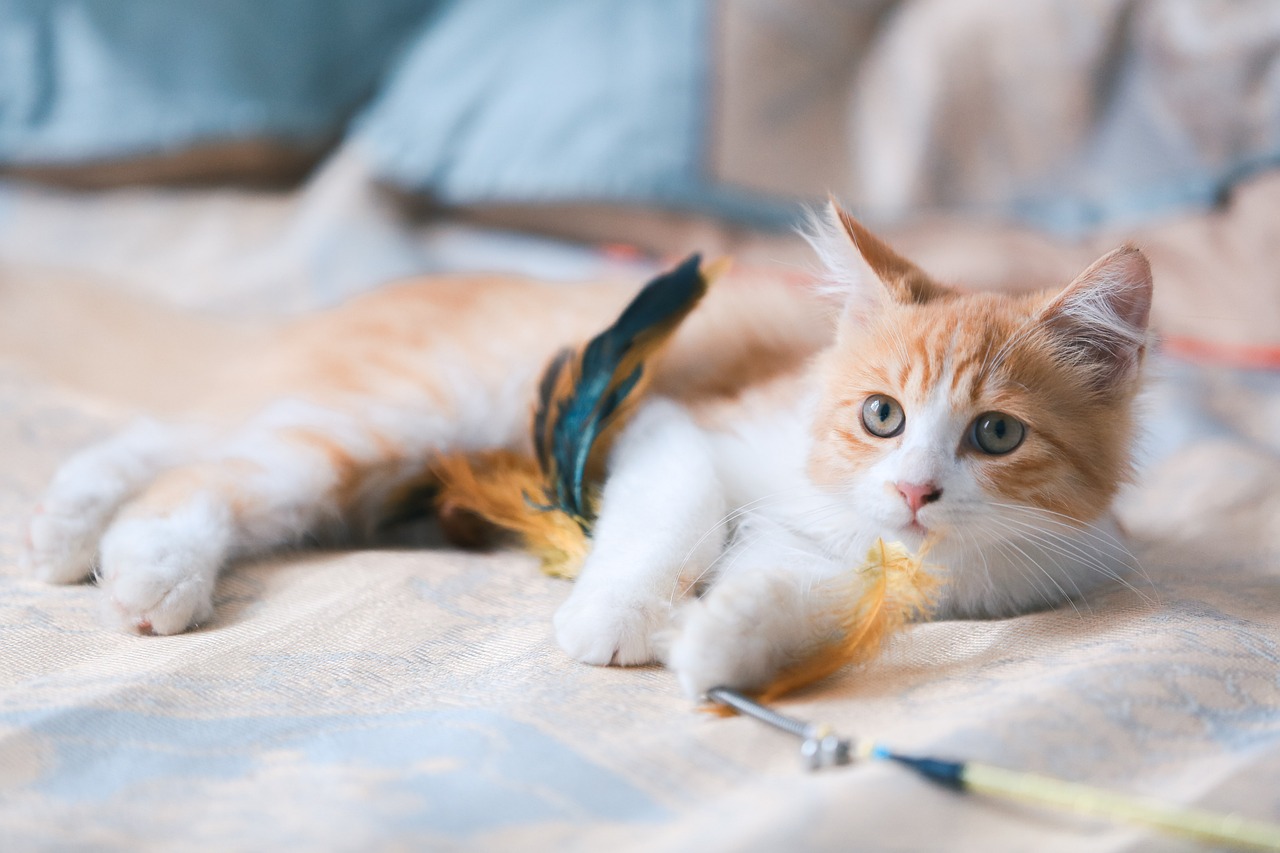}
        \caption{Original}
        \label{fig:origin}
    \end{subfigure}
    \hfill
    \begin{subfigure}[b]{0.11\textwidth}
        \includegraphics[width=1.0\textwidth]{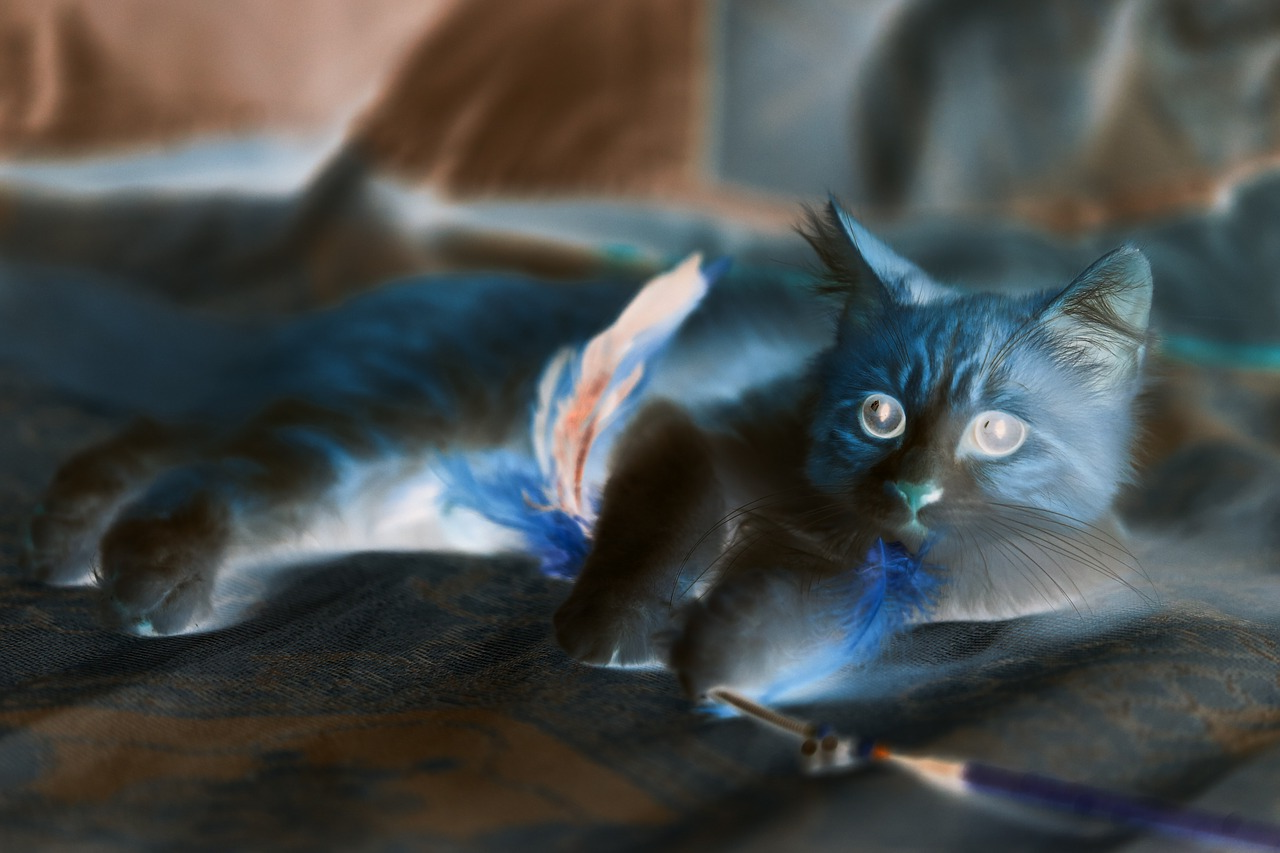}
        \caption{Color}
        \label{fig:color}
    \end{subfigure}
    \hfill
    \begin{subfigure}[b]{0.11\textwidth}
        \includegraphics[width=1.0\textwidth]{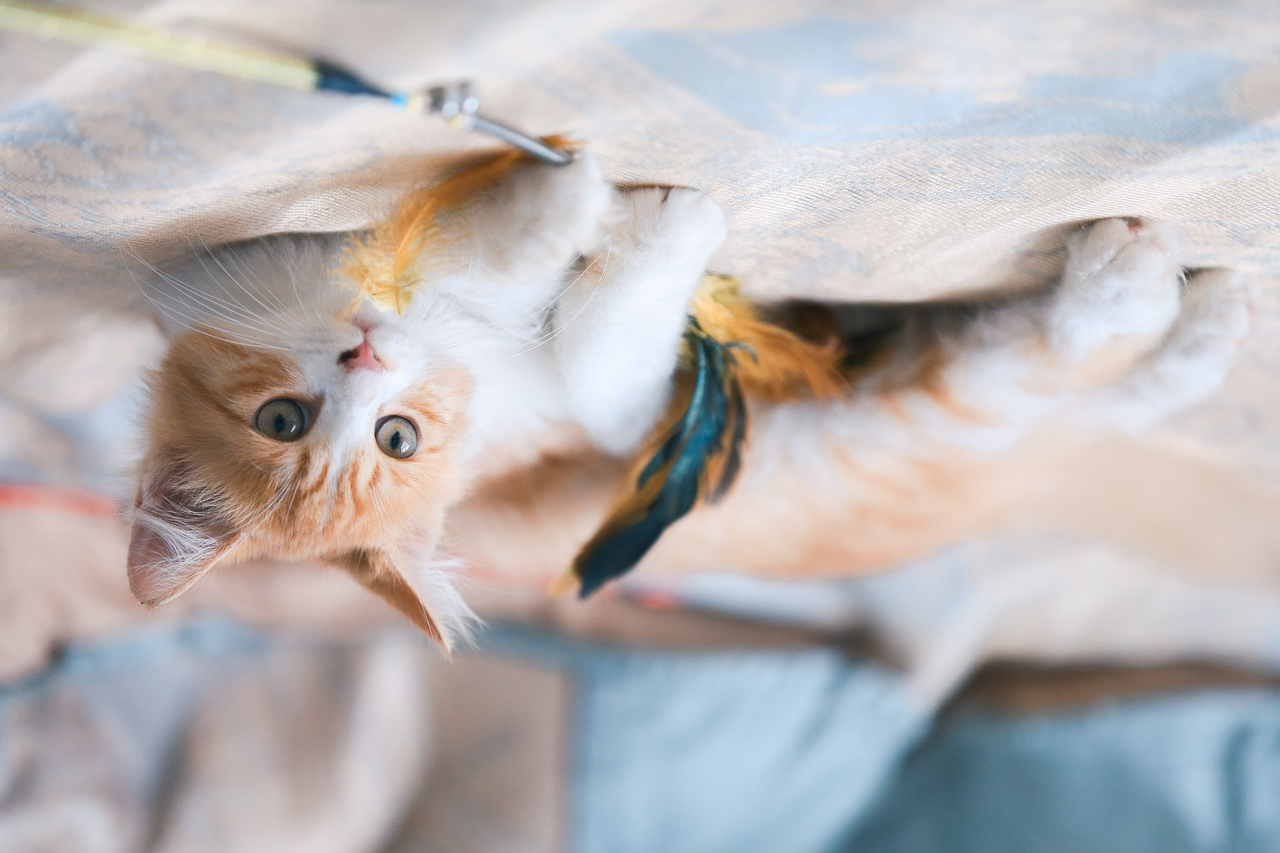}
        \caption{Flip}
        \label{fig:flip}
    \end{subfigure}
    \hfill
    \begin{subfigure}[b]{0.11\textwidth}
        \includegraphics[width=1.0\textwidth]{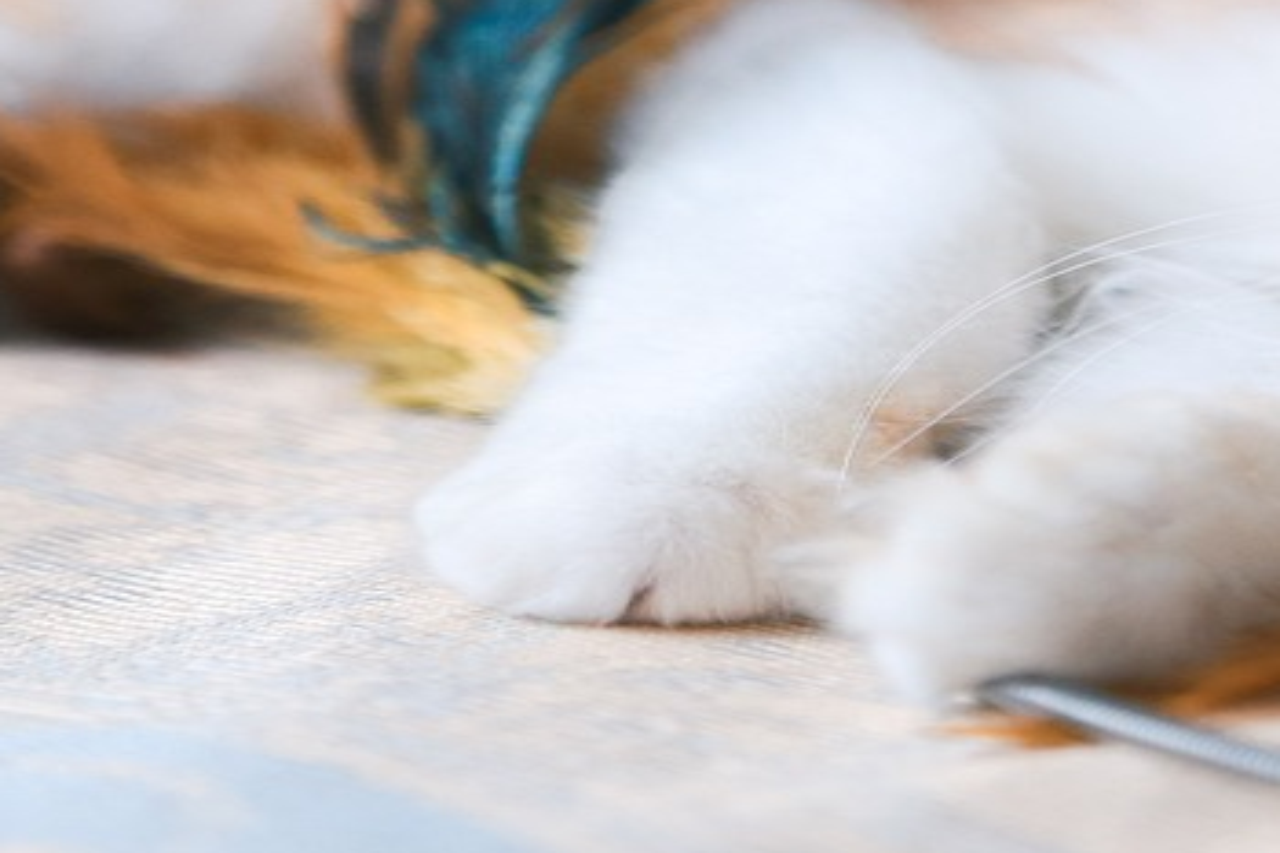}
        \caption{Crop}
        \label{fig:crop}
    \end{subfigure}
    \hfill
    \begin{subfigure}[b]{0.11\textwidth}
        \includegraphics[width=1.0\textwidth]{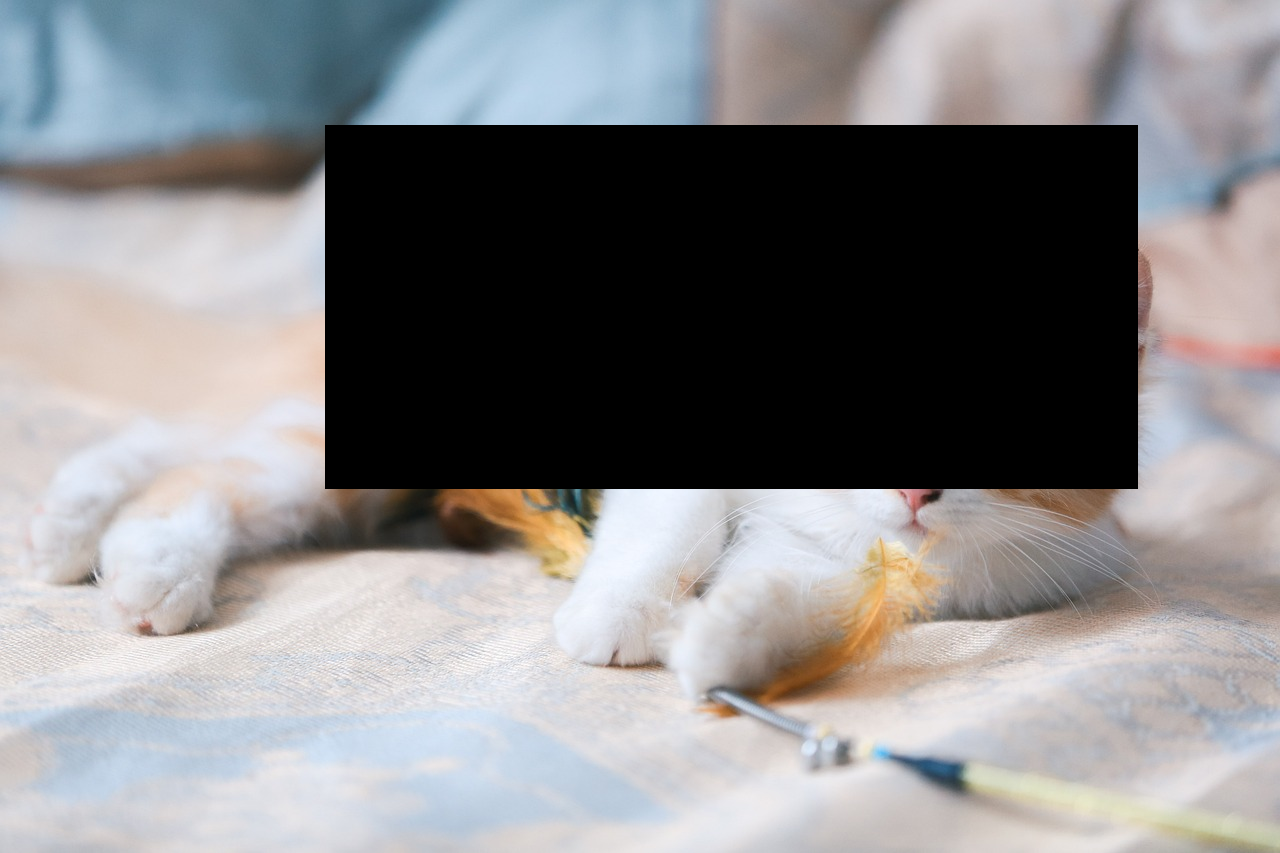}
        \caption{Erase}
        \label{fig:erase}
    \end{subfigure}
    \hfill
    \begin{subfigure}[b]{0.11\textwidth}
        \includegraphics[width=1.0\textwidth]{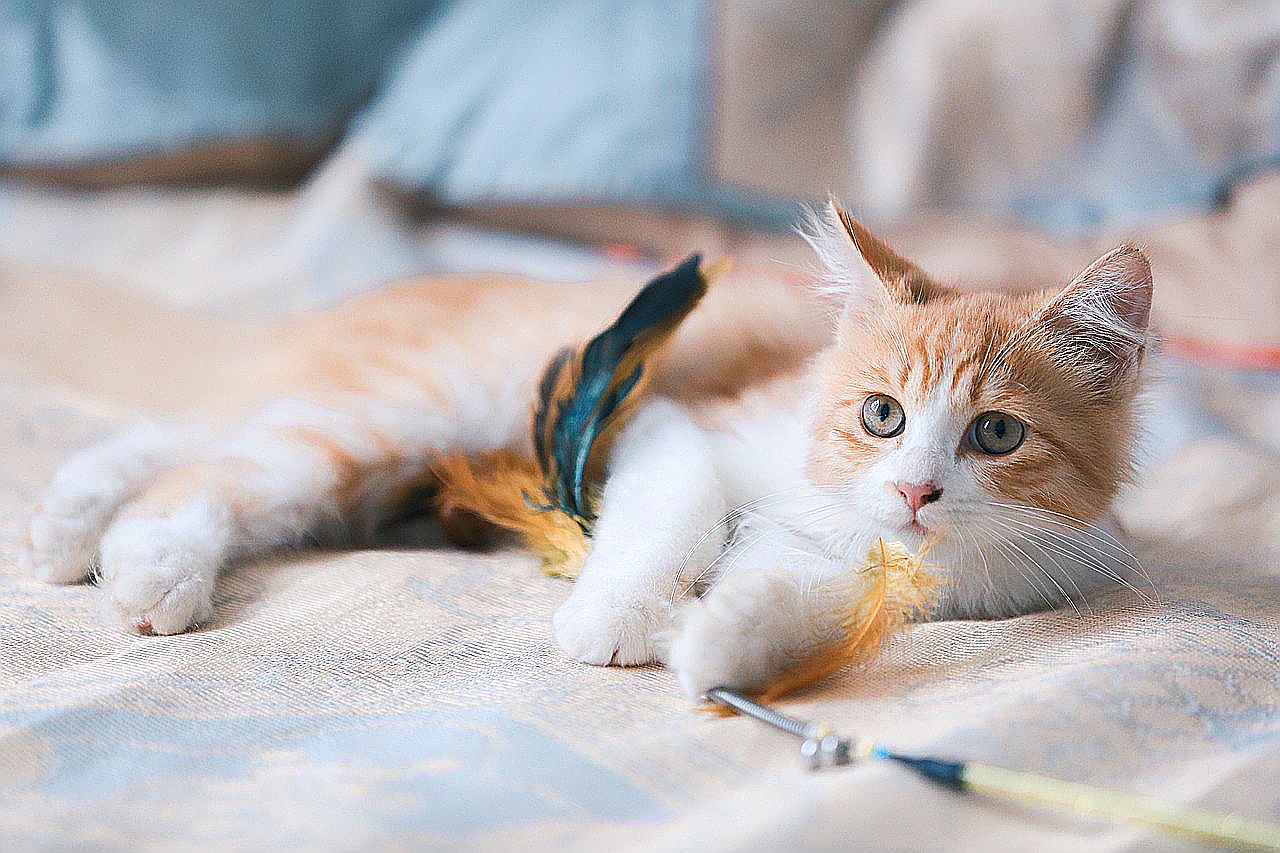}
        \caption{Sharp}
        \label{fig:sharp}
    \end{subfigure}
    \hfill
    \begin{subfigure}[b]{0.11\textwidth}
        \includegraphics[width=1.0\textwidth]{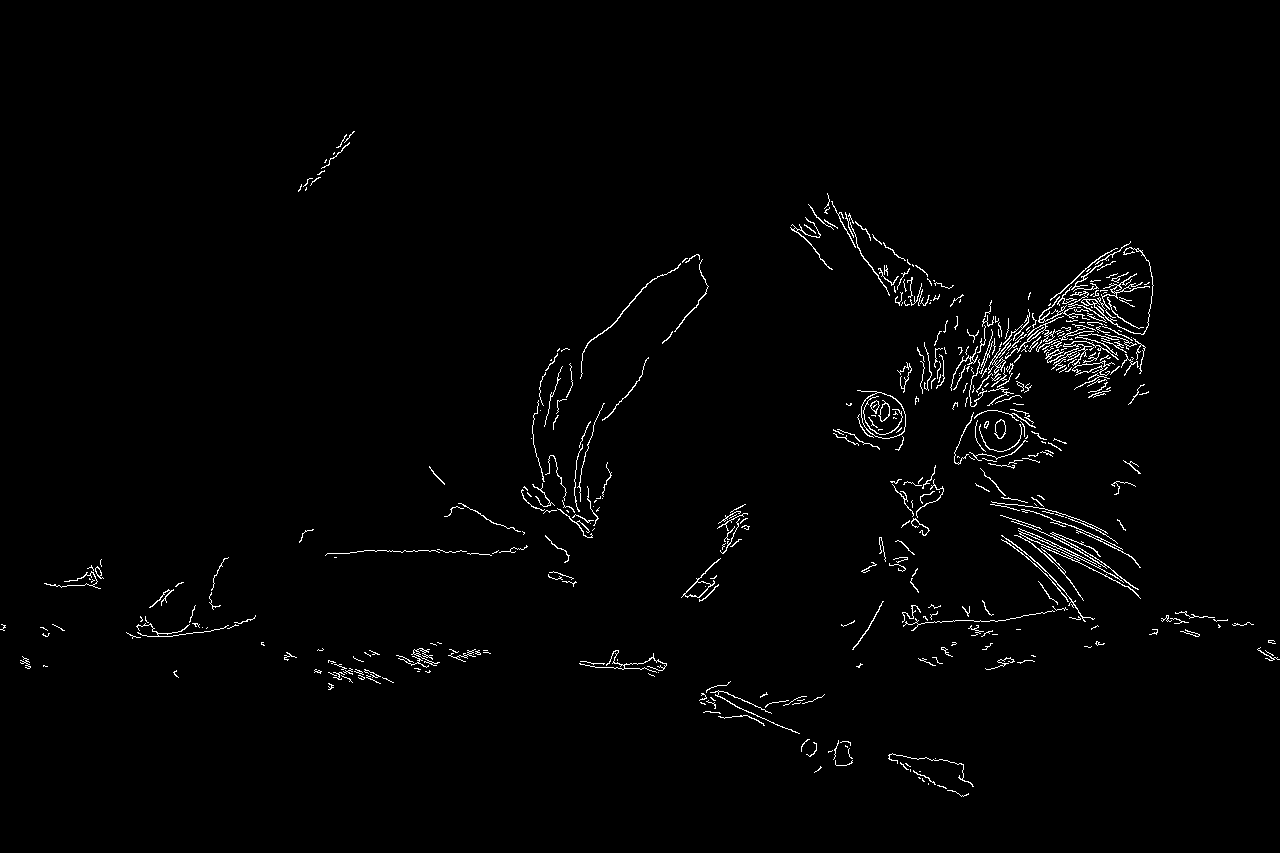}
        \caption{Edge}
        \label{fig:edge}
    \end{subfigure}
    \hfill
    \begin{subfigure}[b]{0.11\textwidth}
        \includegraphics[width=1.0\textwidth]{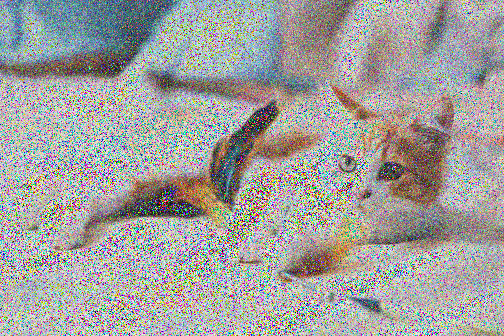}
        \caption{Noise}
        \label{fig:noise}
    \end{subfigure}
    \caption{The examples of visual augmentation outputs utilized in this paper.}
    \label{fig:augmentation}
    \vspace{-5pt}
\end{figure}

\myparagraph{Visual Augmentation (VA).}
VA consists of long-established techniques that modify visual data to produce desired images for computer vision research, such as enhancing sharpness, adjusting color jitters, and more. 
While some augmentation techniques, like mixup~\cite{zhang2017mixup} or CutMix~\cite{yun2019cutmix}, require combining more than two images, our discussion focuses on single-image augmentation operations for simplicity. We focus on a specific framework:
\begin{equation*}
    v' = \mc{O}_{o \in \mc{A}}(v),
\end{equation*}
where $o$ represents an augmentation operation within the set $\mc{A}$. In this paper, we employ the augmentations $\mc{A} = \{\code{color}, \code{flip}, \code{random\,crop}, \code{random\,erase}, \code{sharp}, \code{edge}, \code{noise}\}$. Examples are illustrated in~\autoref{fig:augmentation}. The descriptions of the augmentations are: (1) color: color inversion, (2) flip: horizontal flip followed by vertical flip, (3) crop: cropping a random part of the image, (4) erase: randomly erasing part of the image, (5) sharp: adjusting image sharpness, (6) edge: extracting edge textures, and (7) noise: adding diffusion noise. Note that we use the default noise setting from VCD~\cite{leng2023mitigating}.

\myparagraph{Contrastive Decoding (CD).} CD approach was first introduced in the language domain~\cite{li-etal-2023-contrastive}. It operates by generating two outputs using two different models: an expert model that produces the original outputs and an amateur model that generates contrastive outputs. CD is then performed based on the contradictions between them. This approach has also been explored in the VLM domain by using manipulated images to create contrastiveness. This method involves using the image to remove unrelated information, such as hallucinations, by subtracting the image with amplified contrastive information from the original one. The process operates as follows:

\begin{equation}
\label{eq:cd}
p_{\text{CD}}(y | v, \mc{O}_o, q) = \code{SOFTMAX} \Bigl( (1 + \alpha) f(y | v, q) - \alpha f(y | \mc{O}_o(v), q) \Bigl),
\end{equation}
where $f(\cdot)$ is the output logit obtained from VLM, $\mc{O}_o(v)$ as augmented image and $\alpha$ as hyperparameter for the strength of contrastiveness. To amplify the hallucination inherent in the VLM, VCD~\cite{leng2023mitigating} added noise to the image and CRG~\cite{wan2024contrastive} used object-wise erasing with the provided bounding box labels. Then, they subtracted the logits of the hallucinated image from the logits of the original image. 

%% file: main/03_method.tex

\section{\texorpdfstring{\alg: \underline{V}isual-\underline{A}ugmented \underline{Co}ntrastive \underline{De}coding}{alg: Visual-Augmented Contrastive Decoding}}
\label{sec:alg}

This section explores the impact of VAs on LVLMs, focusing specifically on CD. In essence, we demonstrate that certain VAs cause either contrast or persistence, implying that the output distribution of the augmented image either varies or stays consistent with the distribution of the original image for the given query. Furthermore, we detail our discovery that contrastive augmentation can be identified using the proposed score, which relies on the softmax distance. Building on these insights, we present a novel algorithm named \alg, which leverages both the original and augmented images for CD.

\subsection{Appropriate Augmentations Enhance the Contrast}
\label{sec:proper_aug}
Initially, we investigate the effect of augmentation on LVLM output. We hypothesize that for each query, there is a specific set of VAs, termed the contrastive augmentation set, which causes the LVLM to produce incorrect answers. These VAs essentially alter the input image, resulting in the loss of key features necessary for correctly answering the query. For instance, if the query is about color, augmentations like color inversion can lead to an incorrect response. In this section, we experiment on the MME benchmark~\cite{fu2024mme} to test our hypothesis and present our findings.

\begin{wraptable}[7]{r}{0.3\linewidth}
\vspace{-10pt}
    \caption{Pairs of query type-contrastive augmentation.}
    \label{tab:contrastive_aug}
    \resizebox{0.3\textwidth}{!}{  
    \begin{tabular}{c|c}
        \thickhline
        Query Type  & Contrast. Aug. \\ \hline
        Color       & Color \\
        Existence   & Random Cropping \\
        Position    & Flip \\
        \thickhline
    \end{tabular}}
\end{wraptable}
\myparagraph{Experimental setting.} 
First of all, we provide a summary of the MME dataset~\cite{fu2024mme} and the experimental settings in detail. MME benchmark categorizes question types into $14$ groups, such as color, count, position, existence, and more. We concentrate on questions related to color, existence, and position to thoroughly investigate the influence of VAs. We manually select contrastive augmentations for each query type, based on our hypothesis that they can produce incorrect outputs. Specifically, depending on the query type, we choose the contrastive augmentation pairs as outlined in~\autoref{tab:contrastive_aug}. Note that the remaining two augmentations for each type are considered persistent augmentations. In subsequent experiments, we use these three augmentations and three query types to evaluate the augmentation effect by assessing softmax outputs with LLaVA-1.5.

\begin{figure}[th!]
    \centering
    \begin{subfigure}[b]{0.49\textwidth}
        \includegraphics[width=1.0\textwidth]{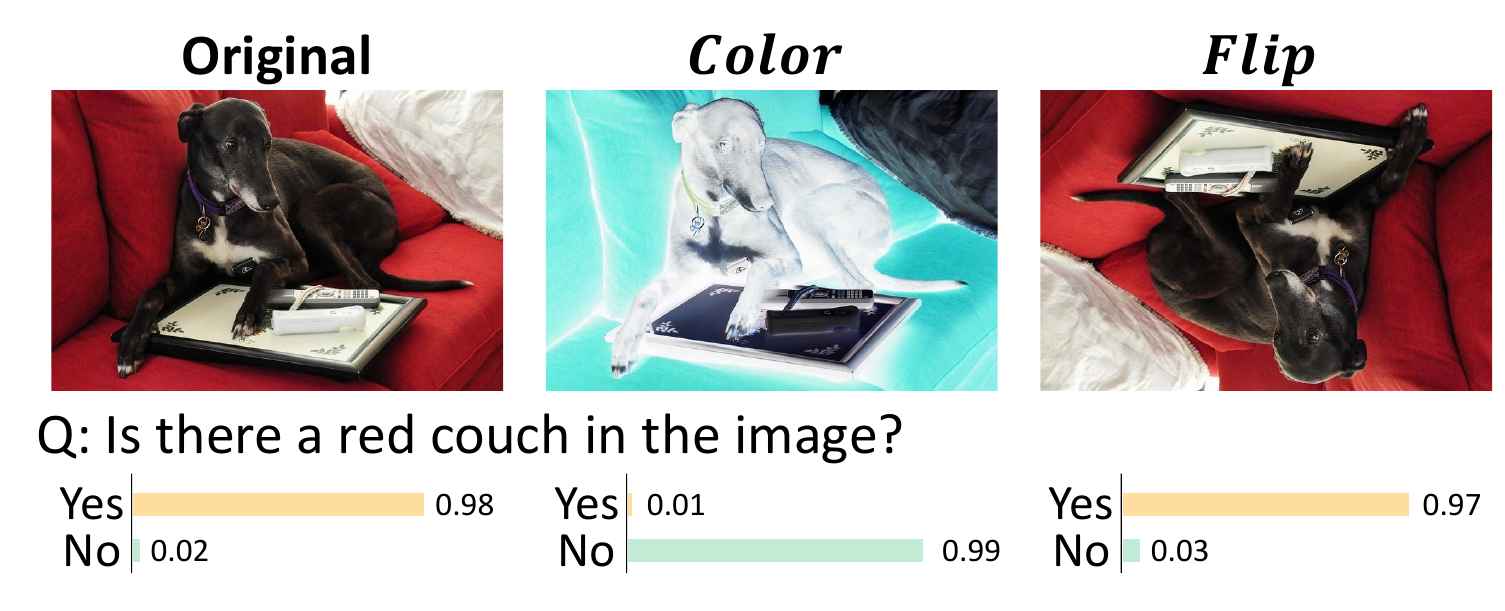}
        \vspace{-20pt}
        \caption{Color-type query}
        \label{fig:mme_color}
    \end{subfigure}
    \hfill
    \begin{subfigure}[b]{0.49\textwidth}
        \includegraphics[width=1.0\textwidth]{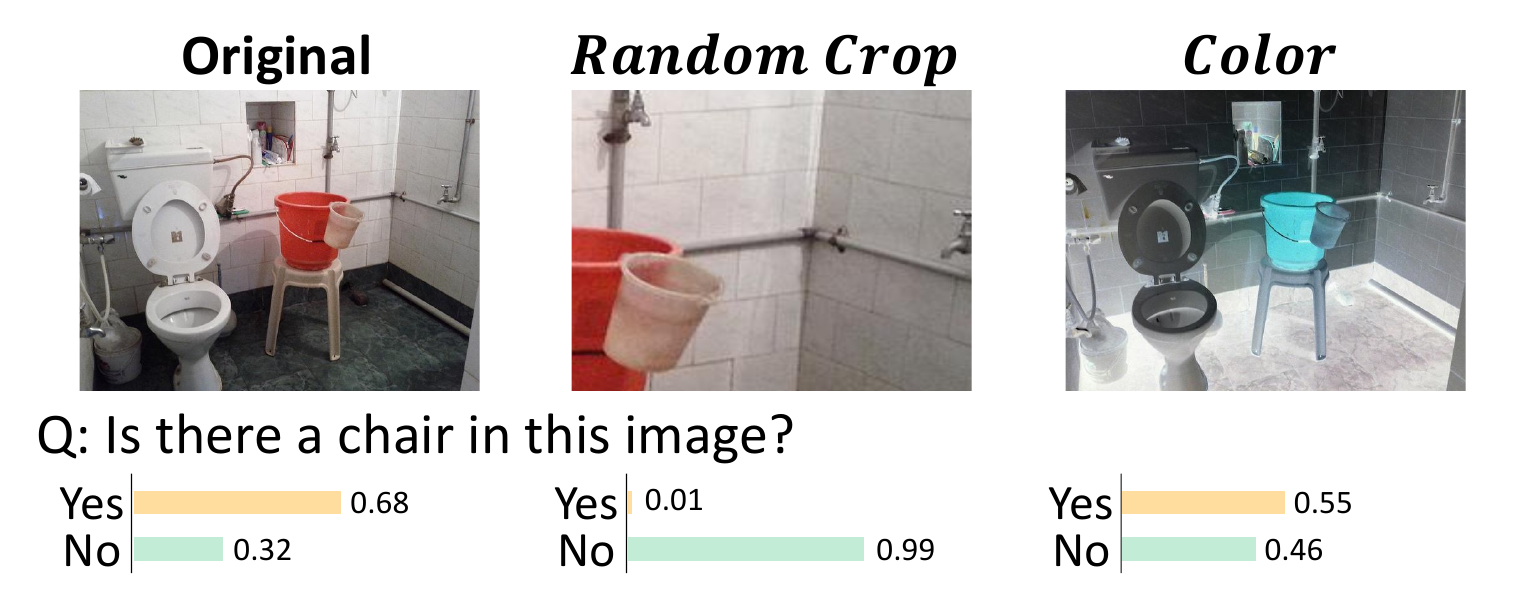}
        \vspace{-20pt}
        \caption{Existence-type query}
        \label{fig:mme_ex}
    \end{subfigure}
    \caption{A detailed analysis of augmentation-question pairs reveals that (a) in color-type query, color augmentation produces a contrastive distribution, whereas flipping does not. Similarly, (b) shows that the existence query is influenced by random cropping.}
    \label{fig:mme_example}
\end{figure}

\begin{figure}[t]
\centering
\begin{minipage}{0.39\textwidth}
    \centering
    \includegraphics[width=0.6\textwidth]{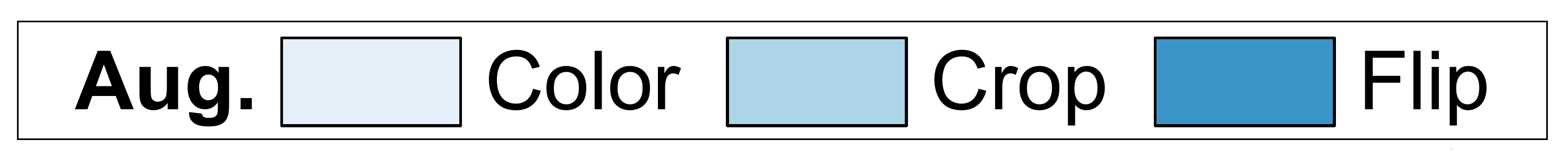}\\        
    \begin{subfigure}[b]{0.49\textwidth}
        \centering
        \includegraphics[width=1\textwidth]{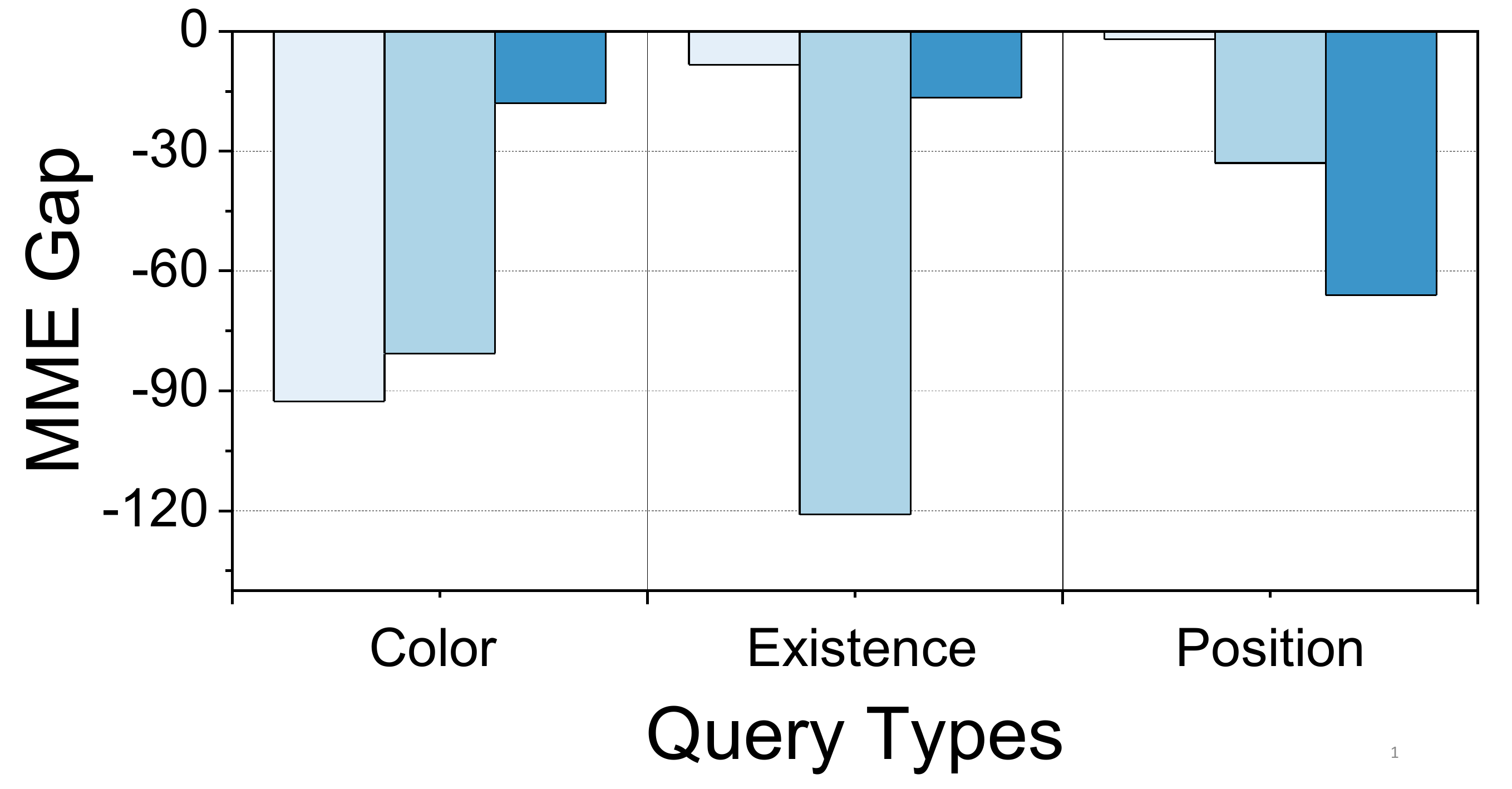}
        \subcaption{MME score drop.}
        \label{fig:mme_heatmap}
    \end{subfigure} 
    \begin{subfigure}[b]{0.49\textwidth}
        \centering
        \includegraphics[width=1\textwidth]{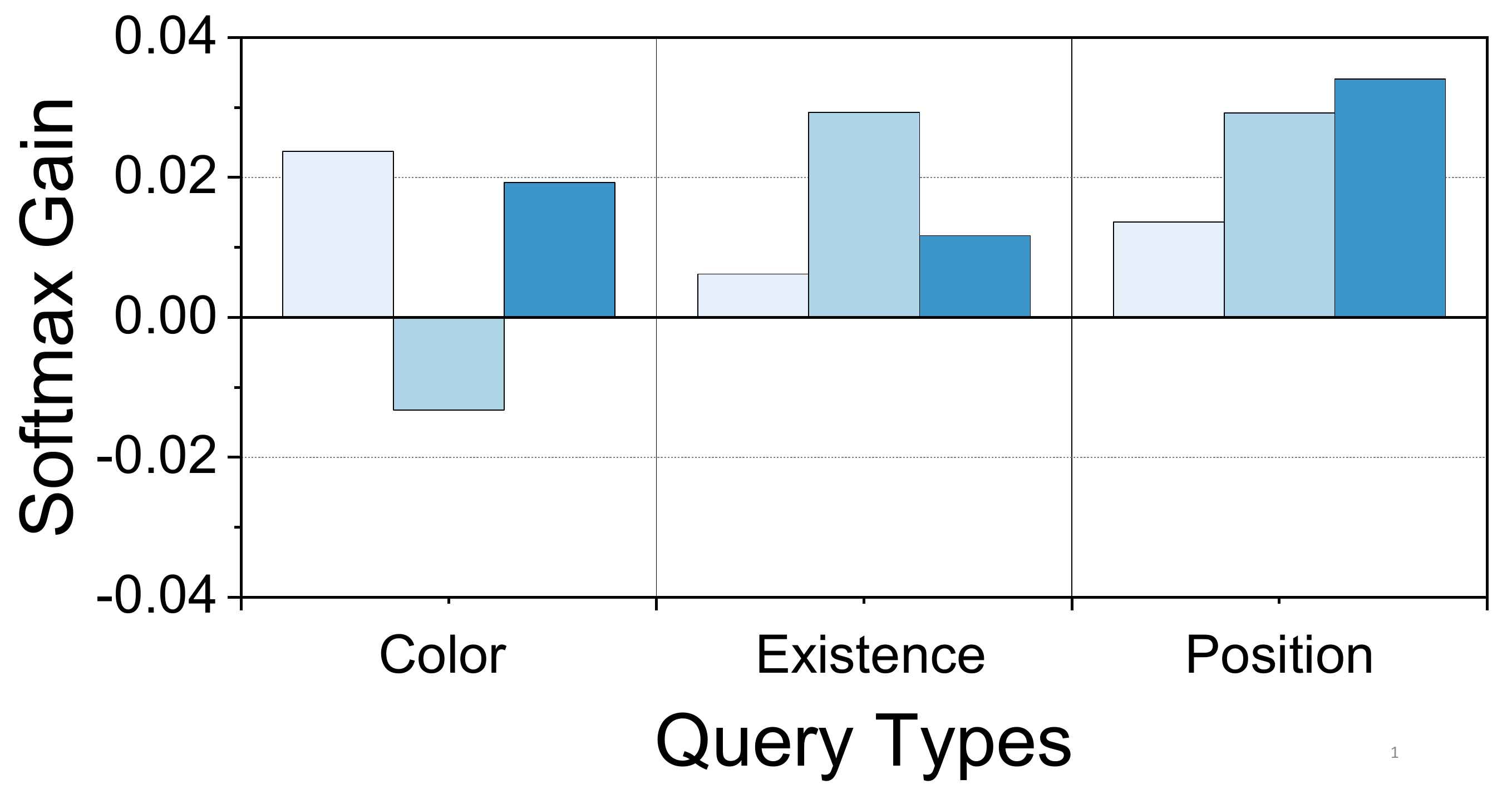}
        \subcaption{Softmax gain.}
        \label{fig:softmax_heatmap}
    \end{subfigure} 
    \caption{On each question type in MME dataset, (a) MME score drop of augmented images and (b) softmax output gain after CD are measured on different augmentations.}
    \label{fig:mme}
\end{minipage}
\hfill
\begin{minipage}{0.29\textwidth}
        \centering
        \includegraphics[width=1\textwidth]{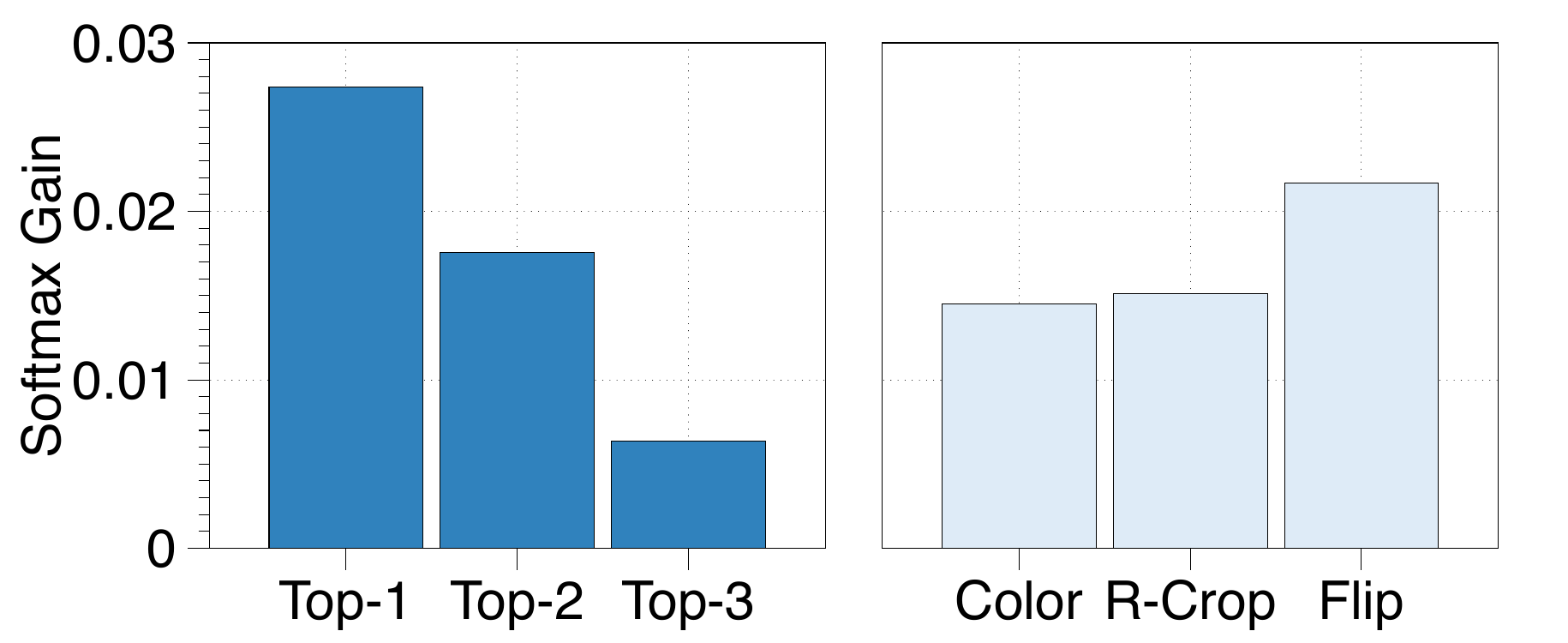}
        \caption{The softmax output of ground truth increases after CD. Top1, the augmentation with the biggest distance, gets the best increment, which is greater than the results of single augmentations. }
        \label{fig:distance_rank}
    \end{minipage}
    \hfill
    \begin{minipage}{0.29\textwidth}
        \centering
        \includegraphics[width=1\textwidth]{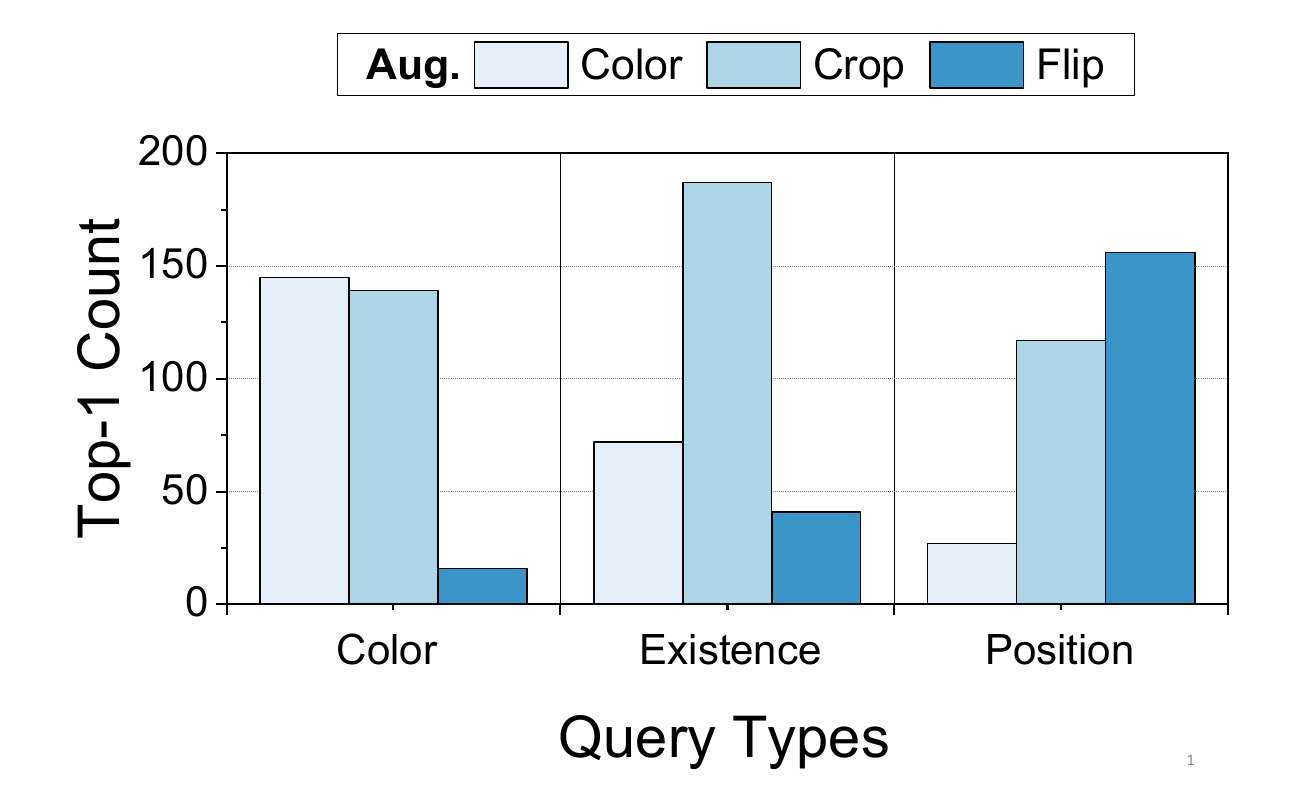}
        \caption{The percentage of augmentations with the largest $D$ in each category shows that, for each category, contrastive augmentation was chosen the most frequently.}
        \label{fig:distance_aug}
    \end{minipage}
\end{figure}

\myparagraph{Contrastive augmentations decrease performance.} 
Using the manually selected query-augmentation pairs, we analyze their outputs both qualitatively and quantitatively. 
For qualitative analysis, as described in~\autoref{fig:mme_example}, we show examples of contrastive and persistent augmentations for given queries. In summary, when the LVLM encounters a contrastively augmented image, it fails to produce the correct answer. 
As shown in~\autoref{fig:mme_color}, when the model is asked a color-related question with the original image, it provides the correct answer. However, when given the color-augmented image, which is a contrastive augmentation on the color-type query, the model generates an incorrect answer. Conversely, if a flip augmentation, \ie persistent augmentation, is applied, the model correctly predicts the answer. This is because flipping does not alter the key color features that are needed to respond accurately. 
Similarly, for the existence-type query, as shown in~\autoref{fig:mme_ex}, random cropping corresponds to a contrastive augmentation since it removes the objects needed to answer correctly. For instance, the portion containing the ``chair'' may be cropped out, leading to an incorrect response. However, coloring the image does not affect the presence of the ``chair,'' allowing the model to provide the correct answer. Qualitatively, there are VAs that can manipulate key features crucial for answering the question correctly.
These results are also reflected in the quantitative results, as shown in~\autoref{fig:mme}. \autoref{fig:mme_heatmap} shows the MME score difference when augmentation is applied. For instance, when the question type is ``existence,'' applying random cropping to the input image lowers the MME score. It means that contrastive augmentation can lead to a performance decrease. 

Additionally, we aim to verify whether this contrastive augmentation can be advantageous in a CD setting. To assess this, we measure the softmax gain, called \code{Gain} score, as follows:
\begin{equation}
    \label{eq:gap}
    \code{Gain}(v,q,y_{\text{GT}},\mc{O}) =  p_{\text{CD}}(y_{\text{GT}}|v,\mc{O},q) - \code{SOFTMAX}(f(y_{\text{GT}}|v,q)).
\end{equation}
This score measures the increase of the softmax on the ground truth value from the original decoding to the CD output to examine how much advantage CD can incur. As illustrated in~\autoref{fig:softmax_heatmap}, utilizing CD methods with contrastive augmentation results in a significant increase in the \code{Gain} score. For instance, in the case of existence-type questions, we observe the highest gain when using contrast augmentation, \ie random crop. Since we rely on impractical information, such as manually selected contrastive augmentations, the remaining challenge is to identify the contrastive augmentation for each query without human intervention.

\subsection{Maximizing Contrast: Selecting Augmentation with the Biggest Distance}
\label{sec:max_dist}

To address the challenge of selecting contrastive augmentation, we first set our intuition that the augmentation resulting in the most different output can serve as a contrastive augmentation. To measure the difference, we use one of the useful metrics, the $L_2$ norm, called distance $D$, defined as follows:
\begin{equation}
\label{eq:distance}
    D(p(v), p(\mc{O}(v))) = \bignorm{\Bigl( p(v) - p(\mc{O}(v)) \Bigr)}_{2}, \mathrm{where} \ \text{$p(v)$} = \code{SOFTMAX} (f(y | v, x) ).
\end{equation}
Note that it can be changed to other types of distance metrics, such as the $L_n$ norm, KL divergence, and so on. Analysis of this metric is also included in~\autoref{app:distance}.

\myparagraph{Choosing augmentation with distance $D$.}
To verify our hypothesis -- \emph{a bigger distance $D$ can have the most contrastiveness} -- we measure distance $D$ following~\autoref{eq:distance} and the \code{Gain} score defined as~\autoref{eq:gap}, under the aforementioned experimental conditions. 
To examine the correlation between the distance $D$ and the \code{Gain} score, we sort the augmentations based on the distance $D$, and analyze the average \code{Gain} score on each ranking.
As shown in~\autoref{fig:distance_rank}, we confirm that the augmentation with the greatest $D$ results in the biggest average increase in the \code{Gain} score. This implies that selecting the augmentation with the highest $D$ yields the best performance improvement. Additionally, the top-ranked group shows a higher increase than other single augmentations. It also indicates that augmentation integration is successful using distance $D$. Moreover,~\autoref{fig:distance_aug} shows how frequently each augmentation is selected as having the highest $D$ score with the knowledge of the question type. The most frequent augmentations correspond to contrastive augmentation, which aligns with intuition in~\autoref{sec:proper_aug}. This suggests using the augmentation with the largest $D$ to select the contrastive augmentation $o$ for each query $q$. While this is an impractical test, it is valuable for understanding the impact of the distance metric.

\input{algorithm/algorithm}

\subsection{\alg: Visual-Augmented Contrastive Decoding}
Based on the above observations, we propose \alg to automatically select an appropriate augmentation for each query by utilizing the distance $D$. 
The entire procedure is summarized in~\autoref{alg:cap}. 
In the initial decoding phase with the given question, we adaptively select contrastive augmentation by calculating the distance metric $D$ and choosing the augmentation with the maximum $D$. This chosen augmentation $\hat{o}$ is then used for the remainder of the decoding process. This approach is adopted because computing the output for all augmentation candidates at every step is too resource-intensive. Once the contrastive augmentation is determined, LVLM calculates the probability of tokens $p_{\alg, t}$ using \autoref{eq:cd}. Subsequently, among the whole vocabulary $V$, the candidate vocabulary set $V_{\text{cand}} \in V$ is defined to select a more reliable token following the original CD algorithm~\cite{li-etal-2023-contrastive}. This process is repeated iteratively to generate the output words $y_t$.

For \alg, we use two scenarios to define candidate augmentations: \emph{all} and \emph{selection}. \emph{All} uses all the augmentations as augmentation candidate set $\mc{A}$. However, some augmentations may be ineffective or replaceable depending on the given query types or the augmentations in the set. In this case, excluding these augmentations may work as a way to eliminate noisy augmentations. So we introduce the \textit{selection} strategy that leverages validation to choose a subset of augmentations $\mc{A'} \in \mc{A}$ to use only the more effective augmentations. Detailed settings and methods are explained in~\autoref{app:selection}.

%% file: algorithm/algorithm.tex
\begin{algorithm}[t]
\caption{\alg: Visual-Augmented Contrastive Decoding}
\label{alg:cap}
\begin{algorithmic}
    \State \textbf{Input}: Image and question pair $(v,q)$, target sequence length $T$, augmentation set $\mc{A}$, distance function $D(\cdot)$, vocabulary set $V$, amplification coefficient $\alpha$, plausibility constraint parameter $\beta$
    
    \For{$t = 1...T$}
        
        \If {$T = 1$} \algcomment{Determine contrast augmentation for the entire decoding process}
            \State $z_{t} \gets f(y_t | v, q, y_{<t})$ and $\tilde{z}_{t, i} \gets f(y_t | \mc{O}_{o}(v), q, y_{<t}), \quad \forall o \in \mc{A}$ \algcomment{Generate logits}
            \State $p_{t} \gets \code{SOFTMAX}(z_{t})$ and $\tilde{p}_{t, i} \gets \code{SOFTMAX}(\tilde{z}_{t, i}), \quad \forall o \in \mc{A}$ \algcomment{Compute probability}
            \State $\hat{o} \gets \argmax_{o \in \mc{A}}(D(p_t, \tilde{p}_{t, i}))$ \algcomment{Select the most constrastive augmentation}
        \Else
            \State $z_{t} \gets f(y_t | v, q, y_{<t})$ and $\tilde{z}_{t, \hat{o}} \gets f(y_t | \mc{O}_{\hat{o}}(v), q, y_{<t})$  \algcomment{Generate logits}
            \State $p_{t} \gets \code{SOFTMAX}(z_{t})$ and $\tilde{p}_{t, \hat{o}} \gets \code{SOFTMAX}(\tilde{z}_{t, \hat{o}})$ \algcomment{Compute probability}
        \EndIf
        \State $p_{\alg, t}$ = $(1+\alpha)\cdot p_{t} - \alpha \cdot \tilde{p}_{t, \hat{o}}$ \algcomment{Compute \alg probability}
        \State $V_{\text{cand}}(y_{<t}) \gets \{y_{t} \in V: p_{t}(y_{t}|v, q, y_{<t}) \geq \beta \, \max_{w} p_{t}(w | v, q, y_{<t})\}$ \algcomment{Candidate vocab. set}
        \State $p_{\alg, t}(y) = 0, $ if $y \notin V_{\text{cand}}(y_{<t})$ \algcomment{Discard not-candidate tokens}
        \State $y_{t} = \code{SAMPLING}_{y}(p_{\alg, t}$) \algcomment{Sampling next token}
    \EndFor
\end{algorithmic}
\end{algorithm}

%% file: main/04_exp.tex
\section{Experiments}
\label{sec:exp}

In this section, we aim to validate the superiority of the proposed algorithm through both qualitative and quantitative analyses.

\subsection{Experimental Settings}
\label{sec:exp_setting}

\myparagraph{Datasets and evaluation metrics.}
We conduct experiments using three datasets: \code{MME}~\cite{fu2024mme}, \code{MMBench}~\cite{liu2024mmbench}, and \code{VQAv2}~\cite{DBLP:conf/cvpr/GoyalKSBP17}. Each dataset consists of image-question pairs to evaluate how well LVLM generates robust and correct answers to various questions. 
\begin{itemize}[leftmargin=15pt]
\item \code{MME} is an LVLM evaluation dataset with granular question categories, including 10 categories from the perception tasks and 4 from the cognition tasks. 
The labels consist of `Yes' or `No,' and performance is measured by MME score, which is derived from accuracy. In this paper, we evaluate the perception category as our method focuses on observation ability.
\item \code{MMBench} is a dataset of image-question pairs from 20 categories to validate how skillfully LVLM performs on various vision-language tasks with given option labels. For evaluation, we incorporate CircularEval which rotates the positions of the possible option labels in a circular manner.
\item \code{VQAv2} is a dataset containing open-ended questions paired with images. %
This allows for a proper evaluation of how expertly the model can utilize the given visual information rather than simply using the learned language priors of the decoder. We randomly select $30,000$ samples from the \code{VQAv2} evaluation dataset to validate our method.
\end{itemize}
For the reliability of the results, we report performance using the average of the results of 5 different seed runs for \code{MME} and \code{MMBench}, and a single run for \code{VQAv2}.

\myparagraph{Models.}
We evaluate the performance of \alg on three baseline LVLM foundation models: \code{LLaVA-1.5}~\cite{liu2023llava}, \code{InstructBLIP}~\cite{dai2024instructblip} and \code{Qwen-VL}~\cite{bai2023qwenvl}. Specifically, we use pretrained \code{LLaVA-1.5} and \code{InstructBLIP} with \code{Vicuna}~\cite{vicuna2023} 13B language decoder, and \code{Qwen-VL} with \code{Qwen} 7B backbone. Ablation studies on model size can be found in~\autoref{app:model_size}.

\myparagraph{CD methods.} We use 7 augmentations in~\autoref{fig:augmentation} for the augmentation set $\mc{A}$. CD with each single augmentation is used as a baseline, and note that a single noise addition augmentation is equivalent to the VCD~\cite{leng2023mitigating} method. We exclude other methods from the baseline that incorporate external models or data.
When applying \alg, we employ both \textit{all} and \textit{selection} strategies. The \textit{selected} augmentations vary depending on the models or datasets. For example, on the \code{MME} benchmark, the LLaVA-13B model utilizes four specific augmentations: color, edge, crop, and flip for \textit{selection}.

\myparagraph{Implementation details.}
For the main experiment, we choose $\alpha=1$ and $\beta=0.1$ for \alg. Additionally, we use $T=1$ and $p=1$ for the sampling strategy, which employs the softmax distribution for the next token generation. Ablation studies on decoding strategies can be found in~\autoref{app:sampling}.

\input{tab/mme_category}

\input{tab/mme_subplot1}

\subsection{Experiment Results}

\myparagraph{Result on each category.}
\autoref{table:mme_category} shows the MME score of CD using different augmentations on the \code{MME} dataset for each perception category using the LLaVA-1.5 13B model. First, looking at the single method, as mentioned in~\autoref{sec:proper_aug}, if each single visual augmentation corresponds to a contrastive augmentation on the given question, it improves CD performance. Although it does not improve the performance in all other question categories, the total MME score increases. This means LVLMs are likely to provide incorrect answers when image augmentations are applied. For instance, in the case of questions about recognizing celebrities or landmarks, humans can answer the corresponding labels even if the color information is distorted. However, when a color-distorted visual image is given, LVLMs fail to perceive the object, and the contrast increases significantly. Through these observations, we can indirectly figure out some impacts of augmentations on the LVLMs. 

Using \alg results in better performance compared to using a single augmentation. As shown in~\autoref{fig:mme} and~\autoref{table:mme_category}, when using a single augmentation for CD of LVLMs, it is challenging to gain distinguished performance across all types of questions. However, \alg automatically selects the candidate expected to have high contrast based on the given task among the candidate augmentations and uses it for CD. Selecting an appropriate visual augmentation based on a given question and image shows outstanding performance improvement across overall question categories compared to using single visual augmentations. 

\myparagraph{Results across more datasets and models.}
\autoref{table:mme_vqav2_total} presents the results for the \code{MME}, \code{VQAv2}, and \code{MMBench} datasets using LLaVA-1.5 13B, Qwen-VL 7B, and InstructBLIP 13B models. Performance on \code{MME} is measured by the MME score, while accuracy is used for the other benchmarks. A notable observation is that \alg shows a significant performance improvement in each setting, regardless of the dataset or model used. This indicates the robustness of \alg in improving the accuracy and reliability of LVLM outputs.

\subsection{Discussion}

\myparagraph{Analysis on the \textit{selection} strategy.}
Among the augmentation strategies \textit{all} and \textit{selection}, \textit{selection} shows better performance in~\autoref{table:mme_category} and~\autoref{table:mme_vqav2_total}. This indicates that our approach to eliminating noisy augmentations is effective and highlights the importance of using only the most effective augmentations to achieve better performance. This approach not only proves its efficacy but also provides users with guidance on choosing the optimal subset of augmentations from various options.

\input{tab/mme_subset}

\myparagraph{Analysis of the combination of visual augmentations.}
In this section, we evaluate different combinations of augmentations to estimate the impact of each augmentation. For simplicity, we limit the augmentation set to $\{\code{color}, \code{flip}, \code{random crop}\}$.
\autoref{table:mme_subset} shows the effect of using all augmentation candidates in the set and the impact of excluding each one individually. According to the results, \alg performance using all three augmentations, color, crop and flip, shows higher performance than other sub-combinations. Specifically, when color or flip augmentation is removed from the augmentation set, performance in the color and position categories significantly decreases. Considering each augmentation has a different contrastive effect, the results confirm that providing an appropriate combination of VAs for $\mc{A}$ can provide proper contrast for a given task.

\begin{figure}[t]
    \begin{subfigure}[b]{0.49\textwidth}
        \centering
        \includegraphics[width=1\textwidth]{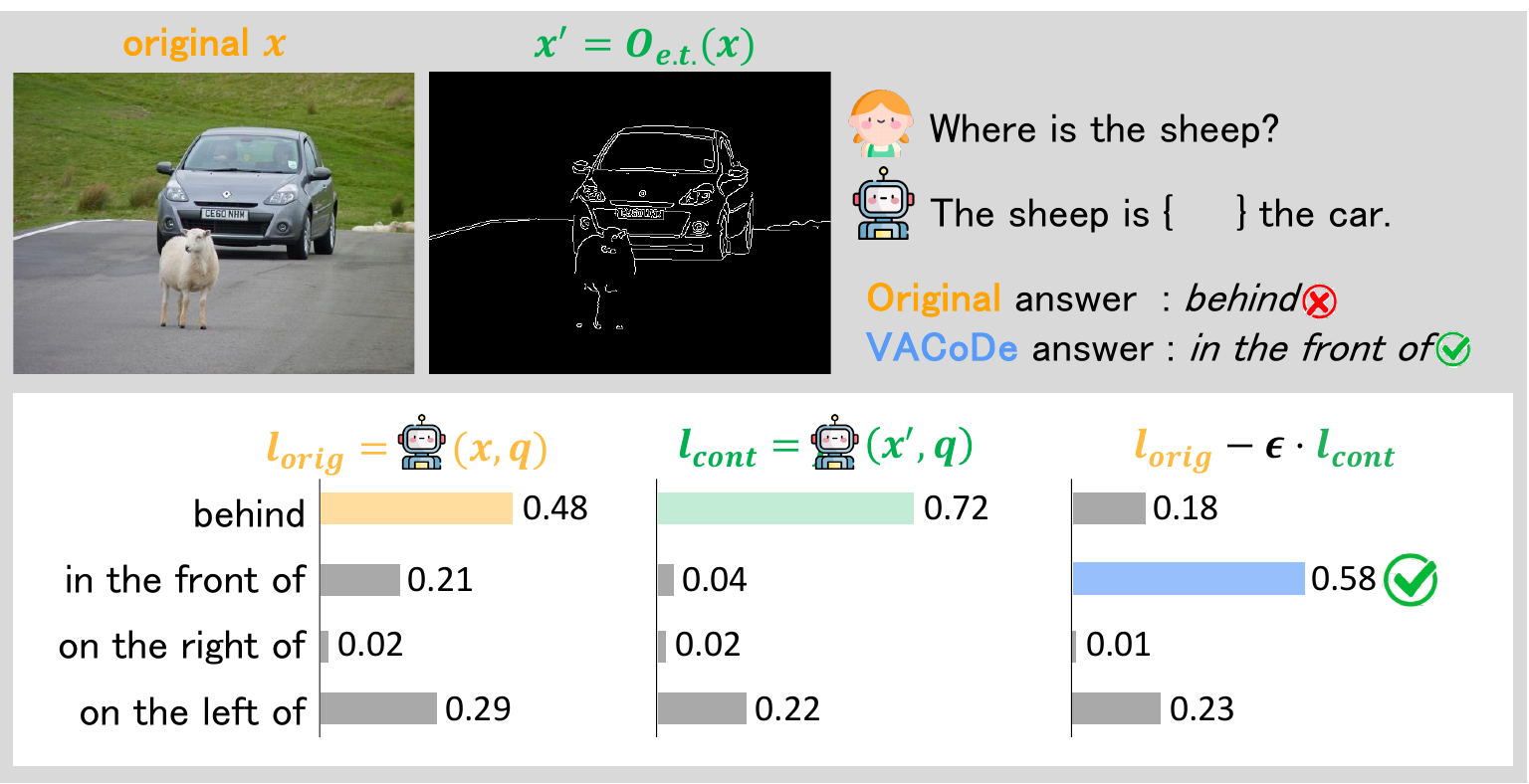}
        \subcaption{Case study 1}
        \label{fig:case_1}
    \end{subfigure} 
    \begin{subfigure}[b]{0.49\textwidth}
        \centering
        \includegraphics[width=1\textwidth]{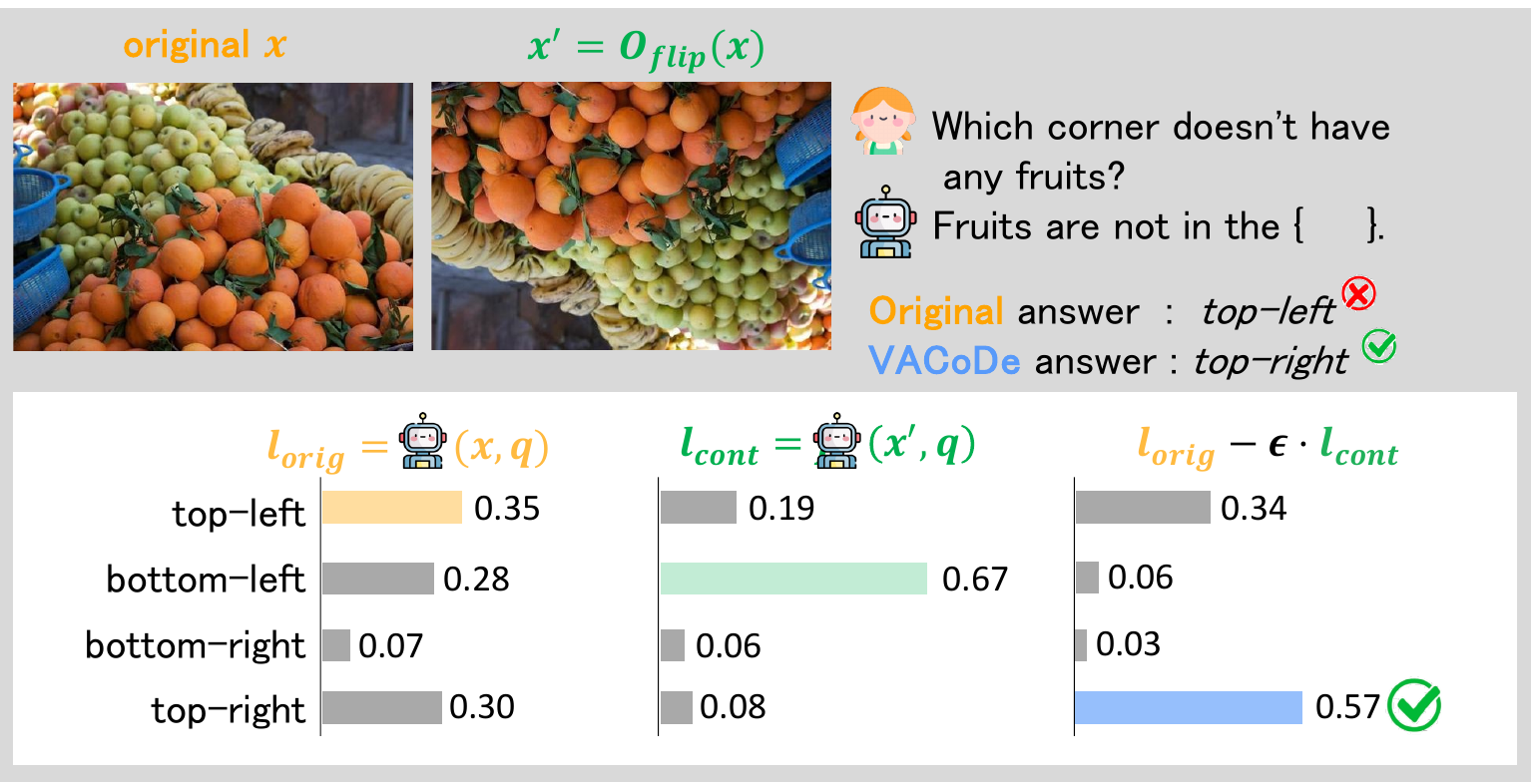}
        \subcaption{Case study 2}
        \label{fig:case_2}
    \end{subfigure} 
    \caption{Two case study examples of \code{MMBench}: in both cases, the original images yield incorrect answers. However, \alg successfully provides correct results. For the contrastive augmentations, \alg selects edge augmentation in the first case and flip augmentation in the second case, both leading to accurate answers.}
    \label{fig:case_study}
    \vspace{-10pt}
\end{figure}

\myparagraph{Qualitative study.}
In this section, we discuss examples of using \alg in \code{MMBench} with LLAVA-13B as illustrated in~\autoref{fig:case_study}. The first example in~\autoref{fig:case_1} demonstrates an instance where LVLM incorrectly predicts  as ``sheep is behind the car.'' When edge augmentation is applied by \alg, it exacerbates LVLM's confusion, increasing the likelihood of an incorrect answer. However, CD corrects this by addressing the disadvantage and generates the correct answer. In~\autoref{fig:case_2}, the original image prediction is `top-left' for an empty space. After flipping the image, the model correctly identifies `bottom-left' as the accurate answer for the augmented image. It's important to note that `top-left' has a high probability in both images, indicating a bias in the LVLM to the given image and the question. Applying CD reduces this bias, resulting in the correct output.

%% file: tab/mme_category.tex
\begin{table*}[!t]
\vspace{-5pt}
\caption{\code{MME} performance on perception task by using LLaVA-1.5 13B. The best and second-best performances are reported using \textbf{bold} and \underline{underline} formatting, respectively.}
\label{table:mme_category}
    \small
    \centering
    \renewcommand{\arraystretch}{1.0}
    \resizebox{\textwidth}{!}{
    \setlength{\tabcolsep}{3pt}{
    \begin{tabular}{l|l|cccccccccc|>{\columncolor{SKY}}c}
    \thickhline
        Method  & Aug. & existence & count & position & color & posters & celebrity & scene & landmark & artwork & OCR & Total  \\  \hline
Regular &   -          & 182.00               & 125.33               & 110.33               & 154.67               & 128.57               & 123.00               & 153.05               & 131.30               & 108.30               & 111.00               & $1327.55_{\pm 16.2}$                             \\ \hline
VCD     & noise & 185.00               & 122.33               & 125.00               & 151.67               & 137.62               & 133.12               & 151.15               & 139.10               & 110.85               & 98.50                & $1354.34_{\pm 24.5}$                           \\ \hline
\multirow{6}{*}{\shortstack{Single}}  & color       & 182.00               & 134.00               & 129.33               & 160.00               & 142.86               & 142.24               & 154.60               & 143.40               & 112.60               & 113.50               & $1414.53_{\pm 9.56}$                            \\
        & edge & 185.00               & 146.00               & 125.00               & 157.67               & 141.70               & 142.24               & 152.95               & 139.50               & 113.15               & 121.00               & $1424.20_{\pm 22.0}$                            \\
        & sharp   & 182.00               & 113.33               & 130.00               & 156.33               & 136.46               & 130.76               & 156.90               & 137.10               & 109.85               & 109.00               & $1361.74_{\pm 20.3}$                             \\
        & crop    & 187.00 & 110.33 & 138.33 & 147.67 & 149.80 & 146.65 & 156.70 & 146.65 & 105.75 & 103.50 & $1392.38_{\pm 24.7}$                             \\
        & erase   & 185.00               & 126.67               & 116.33               & 144.67               & 147.55               & 128.29               & 156.60               & 132.85               & 110.95               & 117.00               & $1365.91_{\pm 22.5}$                            \\
        & flip        & 183.00               & 122.00               & 129.00               & 155.00               & 143.61               & 132.12               & 151.45               & 133.90               & 109.55               & 115.00               & $1374.62_{\pm 14.9}$                        \\
\hline
\multirow{2}{*}{\shortstack{\alg}} & \textit{all}      & 184.00               & 138.67               & 134.00               & 167.00               & 146.80               & 144.29               & 149.35               & 145.30               & 114.65               & 119.00               & \underline{1443.06}$_{\pm 6.80}$                      \\
    & \textit{selection} & 183.00 & 140.33 & 132.00 & 165.33 & 146.46 & 143.71 & 149.80 & 145.05 & 114.45 & 123.00 & \textbf{1443.14$_{\pm 9.99}$}  \\
        
        \thickhline
    \end{tabular}%
    }
}
\vspace{-5pt}
\end{table*}

%% file: tab/mme_subplot1.tex
\begin{table*}[t]
    \caption{\code{MME}, \code{VQAv2}, and \code{MMBench} performance on different LVLMs. LV, QV, and IB denote the LLaVA-1.5 13B, Qwen-VL 7B, and InstructBLIP 13B, respectively.} %
    \label{table:mme_vqav2_total}
    \centering
    \tiny
    \resizebox{\textwidth}{!}{%
        \begin{tabular}{c|l|ccc|ccc|ccc}
        \thickhline
        \multirow{2}{*}{Method} & \multirow{2}{*}{Aug.} & \multicolumn{3}{c|}{\code{MME}} & \multicolumn{3}{c}{\code{VQAv2}} & \multicolumn{3}{c}{\code{MMBench}} \\
         & & LV & QV & IB & LV & QV & IB & LV & QV & IB \\
        \hline
        Regular & - & 1327.55 & 1355.32 & 1151.45 & 67.54 & 75.38 & 61.82 & 73.74  & 64.49  & 43.75 \\
        \hline
        VCD & noise & 1354.34 & 1406.15 & 1208.44 & 71.29 & 75.54 & 66.64 & 74.55 & 68.53 & 48.80 \\
        \hline
        \multirow{6}{*}{\shortstack{Single}} & color      & 1414.53 & 1422.69 & 1237.71 & 71.94 & \underline{76.26} & 67.26 & 75.42 & 68.95 & 48.06\\
         & edge       & 1424.20 & 1393.32 & 1220.63 & 71.88 & 75.92 & 67.51 & 74.77 & 69.07 & 49.76 \\
         & sharp      & 1361.74 & 1395.14 & 1164.32 & 71.35 & 76.18 & 66.45 & 74.69 & 68.17 & 47.12 \\
         & crop       & 1392.38 & 1396.83 & 1205.55 & 71.22 & 76.06 & 66.03 & 74.67 & 68.33 & 47.59\\
         & erase      & 1365.91 & 1385.33 & 1185.32 & 71.66 & 76.10 & 66.64 & 74.86 & 68.36 & 47.17 \\
         & flip       & 1374.62 & \underline{1425.20} & 1213.61 & 71.76 & 75.54 & 66.81 & 75.34 & 69.38 & 48.69\\
        \hline
        \multirow{2}{*}{\shortstack{\alg}} 
        & \textit{all} & \underline{1443.06} & 1406.05 & \underline{1248.30} & \textbf{72.53} & 76.06 & \underline{67.97} & \underline{75.49} & \underline{69.89} & \underline{50.49}\\
        & \textit{selection}  & \textbf{1443.14} & \textbf{1426.43} & \textbf{1256.09} & \underline{72.46} & \textbf{76.29} & \textbf{67.99} & \textbf{75.57}  & \textbf{70.01} & \textbf{50.67}\\
        \thickhline
        \end{tabular}
        }
        
\end{table*}

%% file: tab/mme_subset.tex
\begin{table}[!t]
\caption{MME performance of \alg with different candidate combinations. We evaluate the performance of the candidate set $\mc{A}$ by excluding each candidate one by one.}
\label{table:mme_subset}
    \small
    \centering
    \renewcommand{\arraystretch}{1.1}
    \resizebox{\textwidth}{!}{
    \setlength{\tabcolsep}{3pt}{
    \begin{tabular}{l|l|cccccccccc|>{\columncolor{SKY}}cc}
    \thickhline
        Method  & Aug & existence & count & position & color & posters & celebrity & scene & landmark & artwork & OCR & Total \\
        \hline
\multirow{1}{*}{\shortstack{Regular}}  
& -           & 182.00 & 125.33 & 110.33 & 154.67 & 128.57 & 123.00 & 153.05 & 131.30 & 108.30 & 111.00 & 1327.55 \\
\hline
\multirow{3}{*}{\shortstack{Single}}  
& color           & 182.00 & 134.00 & 129.33 & 160.00 & 142.86 & 142.24 & 154.60 & 143.40 & 112.60 & 113.50 & 1414.53 \\
        
& crop            & 187.00 & 110.33 & 138.33 & 147.67 & 149.80 & 146.65 & 156.70 & 146.65 & 105.75 & 103.50 & 1392.38 \\
& flip            & 183.00 & 122.00 & 129.00 & 155.00 & 143.61 & 132.12 & 151.45 & 133.90 & 109.55 & 115.00 & 1374.62 \\
\hline
\multirow{3}{*}{\shortstack{VACoDe \\ (subset)}}  
                
& color+crop      & 186.00 & 116.67 & 132.33 & 160.00 & 150.27 & 149.82 & 155.70 & 153.35 & 108.75 & 108.00 & \underline{1420.90}                     \\
 & color+flip      & 181.00 & 138.33 & 136.33 & 161.67 & 145.10 & 141.41 & 150.10 & 141.00 & 113.55 & 108.00 & 1416.50                     \\
 & crop+flip       & 184.00 & 116.00 & 133.33 & 150.67 & 148.57 & 147.94 & 155.55 & 151.50 & 107.80 & 103.50 & 1398.86                     \\
\hline
VACoDe & color+crop+flip & 183.00 & 120.33 & 133.33 & 161.00 & 150.07 & 149.94 & 155.70 & 155.70 & 109.20 & 108.00 & \textbf{1426.28}                      \\
        \thickhline
    \end{tabular}%
    }
}
\vspace{-10pt}
\end{table}

%% file: main/05_related.tex
\section{Related Works}
\label{sec:related}

\myparagraph{Visual augmentation.}
In the computer vision domain, visual augmentation has been employed to increase the diversity of sample data, thereby helping to overcome the challenges associated with acquiring large training datasets and mitigating overfitting issues in environments with limited samples. 
Traditional augmentations include changes in color, cropping, and flipping. Additionally, there are more advanced techniques such as erasing~\cite{kumar2017hide, devries2017improved, zhong2020random}, and other techniques such as mixup~\cite{zhang2017mixup} and CutMix~\cite{yun2019cutmix}. Furthermore, the automatic application of multiple augmentations has been explored~\citep{cubuk2019autoaugment,lim2019fast}.

Some studies in LVLMs employ VA to achieve the desired output in various methods. FGVP~\cite{yang2024fine} adds blur to the background of the image, leaving the main object clear to emphasize it. To focus on each object in the image, ~\citep{chen2023see,surismenon2023vipergpt,lin2024fine} use multiple cropped images, each focusing on a single object to generate the desired output, while~\cite{kim2023exposing} uses inpainting to erase objects to measure the correlation between objects.

\myparagraph{Contrastive decoding.}
CD~\cite{li-etal-2023-contrastive} was introduced in the NLP domain using two differently sized language models. It leverages contrastive output by subtracting the small model's probability from the larger model's to retain the strengths of the large model while eliminating the weaknesses that are evident in the small model. There are variants like DOLA~\citep{chuang2023dola} which utilizes contrast in layer-level outputs and Instructive Decoding~\cite{instructivedecoding} uses two contrastive instructions to generate an output opposite to the original output.

Recently, similar approaches have been applied in LVLMs, utilizing contrastive images to guide the model in generating accurate text, mainly focusing on reducing hallucination in LVLM~\citep{li2023evaluating,liu2023mitigating,tong2024eyes}. VCD~\cite{leng2023mitigating} demonstrates that adding noise to the image can elevate the hallucination inherent in LVLMs, subsequently applying CD to manage the hallucination. Another work CRG~\cite{wan2024contrastive} employs a black bounding box from external data to conceal the object relevant to the question, amplifying hallucination, while HALC~\cite{chen2024halc} uses multiple different cropped images from the detection model and explores multiple pairs of cropped images to find pairs that amplify the information in the cropped image. These works address methods to manage hallucination in LVLMs using a single type of augmentation, which has limitations in generating enough contrast for various types of questions.
Unlike previous studies, \alg explores multiple augmentations and selects the most effective one to answer the question. Moreover, it does not require additional training or an external model, providing direct perturbation to the image.

%% file: main/06_conclusion.tex
\section{Conclusion}
\label{sec:conclusion}

In this paper, we introduce \alg for utilizing multiple augmentations by adaptively choosing contrastive decoding. Initially, we examined the effects of various augmentations and found that their effectiveness depends on the type of question. Specifically, each query has key features that act as clues for answers, and contrastive augmentations can modify these features. Therefore, selecting the contrastive augmentation that creates a significant contrast is essential for improving CD. Based on this, we propose an algorithm called \alg, which selects augmentation by the largest distance $D$. Experiments show that \alg outperforms other methods across different datasets and underscores the importance of selecting appropriate augmentations.

\myparagraph{Limitation.}
Our method selects the appropriate contrastive augmentation among augmentation candidates. No matter how well \alg works and the appropriate augmentation is selected for the given task, if there is no sufficient contrastive augmentation for the task among the candidates, it is difficult to expect a significant performance gain.

\myparagraph{Future work.}
Future work includes implementing an automatic search for candidate augmentation sets suitable for the target task. Additionally, investigating the relationship between visual contrast and language contrast in LVLMs suggests a further direction for expanding this study.

\section*{Acknowledgement}
This work was supported by Institute of Information \& communications Technology Planning \& Evaluation (IITP) grant funded by the Korea government (MSIT) (No.2019-0-00075, Artificial Intelligence Graduate School Program (KAIST), 10\%) and the Institute of Information \& communications Technology Planning \& Evaluation (IITP) grant funded by the Korea government (MSIT) (No. 2022-0-00871, Development of AI Autonomy and Knowledge Enhancement for AI Agent Collaboration, 90\%).

%% file: main/appendix.tex
\appendix
\supptitle

\section{\texorpdfstring{Ablation on Distance metric $D$}{Ablation on Distance metric D}}
\label{app:distance}

In this section, we examine comprehensive several additional ablation experiments that are considerable in the environment in which \alg is applied. Based on these ablation results, we expect \alg to have universally high robustness and be able to perform various tasks, models, and inferences.

We perform experiments using several common distance measures to define our distance function $D$ that \alg uses to select which VA will produce high contrast. The experiment is performed in the MME dataset using the \code{LLaVA-1.5} 13B model. Also, we use the average softmax \code{Gain} directly to check the effect. In detail, softmax \code{Gain} on the correct answer label obtained when applying the distance measure candidate $D_{i}$ used in the experiment and the VAs used in~\autoref{fig:augmentation} to \alg for all samples. In order to control the variables of VAs that contain randomness, each experiment performs a total of 5 experiments with different seeds on the entire \code{MME} dataset and then measures softmax \code{Gain} through the average.

\begin{figure}[h]
    \centering
    \includegraphics[width=0.3\linewidth]{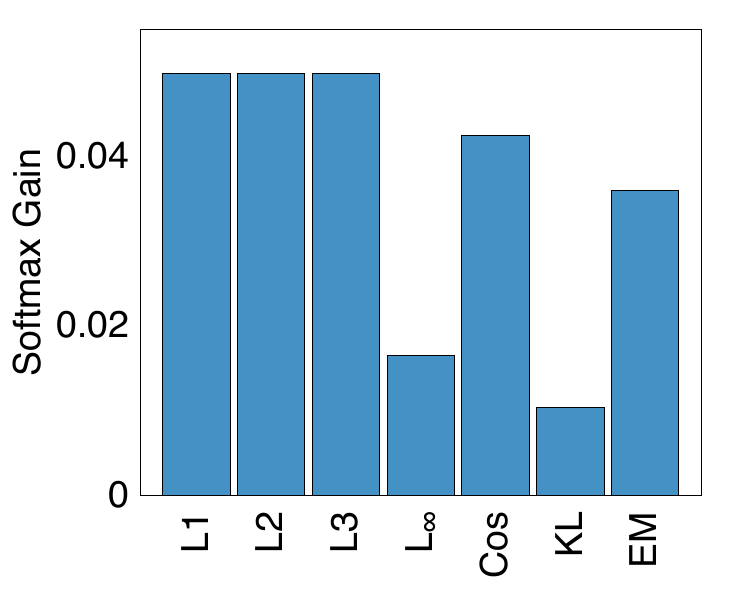}
    \caption{Average softmax gain by different distance metrics.}
    \label{fig:distance_metric}
\end{figure}

\autoref{fig:distance_metric} shows the result of affectness of different distance $D$ functions. In this experiment, we use $L_{1}$, $L_{2}$, $L_{3}$, $L_{\infty}$, Cosine similarity, Kullback-Leibler Divergence (KL divergence), and Earth mover's distance (EM distance) as distance candidates. The x-axis of the results in~\autoref{fig:distance_metric} means the candidate distance names used, and the y-axis means the average softmax gain improved compared to regular decoding obtained through \alg when each distance is used as a measurement. From the results, we can check that $L_{1}$, $L_{2}$, and $L_{3}$ norms show high performance improvement almost no difference overall. This means that any of these can be used in the algorithm as a distance function at a similar level. However, in the case of $L_{\infty}$ and KL divergence, it can be seen that the actual performance improvement is much smaller compared to others. These show very low-performance improvement compared to the $L_{2}$ distance, which we used in the main experiment, meaning they are improper measurements for estimating the expected contrast of VAs. The other two distances, cosine similarity and EM distance, performed higher than KL divergence but did not perform higher than $L_{2}$ norm for the entire \code{MME} dataset. Based on this result, we empirically confirmed that using $L_{2}$ norm as our main \alg distance $D$ is a meaningful standard through experiments with these distance measures and the results shown throughout our main experiments.

\section{Analysis of Different Model Sizes}
\label{app:model_size}

We showed that \alg is proper for general LVLMs and has a significant effect on performance by experimenting with three different models \code{LLaVA-1.5}, {InstructBLIP}, and {Qwen-VL} on various types of datasets at the~\autoref{sec:exp}. In this ablation, we conduct an experiment using \code{LLaVA-1.5} 7B, 13B and {InstructBLIP} 7B, 13B to check the effect of the model size on \alg. \code{MME} dataset is used for this experiment. We measured the performance for the perception category and the total performance for each model.

\input{tab/mme_modelsize}

\autoref{table:mme_modelsize} shows the performance of \alg on each model and size for the \code{MME} dataset. From the result, we can confirm even if the model size and model used are different, the softmax gain obtained when each VA is used in \alg is robust to the type and size of the model and shows a tendency to be dependent on the given task. Throughout the experimental results, the single VA \code{edge} and \code{color} show very high performance. On the other hand, we can see that single VA \code{sharp} and \code{erase} have an overall low-performance gain. For different models, the performance gain shown by each VA shows an overall similar trend, and it can be seen that there is a higher performance improvement compared to the original regular decoding.

Furthermore, for different model sizes, we can see that there is a significant performance gain when applying our algorithm \alg. \alg using all of the VAs specified in~\autoref{fig:augmentation} shows a higher performance improvement than using each single VA. This indicates that, regardless of model and size, each application has the highest performance in the entire perception category and total performance.

\section{Effect of Different Sampling Strategies}
\label{app:sampling}

We perform analysis studies on different sampling strategies to see how \alg is affected by sampling methods other than basic regular decoding. In this experiment, 4 sampling techniques are applied: (1) Top P sampling (specifically, $p=0.9$), (2) Top K sampling (specifically, $k=50$), (3) Temperature sampling (specifically, $T=0.7/1.5$). Top P sampling is a method in which the only token candidates in the distribution on cumulative probability $p$ can be selected as the next token. This has the effect of preventing noise samples with too low a probability to be extracted from candidates. Top K sampling uses only the top $k$ candidates from the highest probability for sampling. In temperature sampling, temperature scaling is applied to the softmax to calculate the next token logits. When temperature $T$ is low, the possibility of selecting a high-probability candidate group increases, and the possibility of choosing low-probability candidates decreases. It has the effect of increasing the probability of more static responses. Conversely, when the temperature $T$ is large, the chance of choosing among the high-probability candidates decreases, and the low-probability candidates increases. It has the effect of increasing the possibility of making more diverse responses.

\input{tab/mme_sampling}

\autoref{table:mme_sampling} show the experiment result of \alg with different sampling strategies. From the table, we can check that \alg gives us a high performance in various types of samplings. This is not only for regular decoding, but it also shows higher performance compared to single VA in the Top P sampling and Top K sampling. A notable observation is that \alg shows high performance in both cases where the temperature scale gets higher or lower. In the case of high temperature, the model has a higher probability of generation more diverse, and the explanations and representations are getting richer. However, in this case, there is a potential problem that the entire output is inaccurate while in generation. In particular, if specific information for a given image must be utilized rather than using inherent prior knowledge, however, there is a possibility that incorrect output may lose correlation with visual information on LVLMs. Our results show that using \alg in this situation can be expected to have the effect of concentrating the model to intentionally utilize visual information by contrastive decoding the output through contrast VA. As can be seen from the results, in situations where the temperature scale is large, CD through VA produces a more significant performance gain. Additionally,  the magnitude of contrastiveness produced by each VA is different in the task so that we can see a considerable performance difference between single VA CDs. In this situation, \alg, which automatically selects and applies the appropriate VA for a given task, can be used more appropriately and robustly to the given scenario. Furthermore, it shows that \alg has the highest performance improvement. \\

Our algorithm can be also used at the low temperature scale scinarios, which grows the sampling possibility of high probabiltiy token being chosen as next token. In this scenario, the original model's high logits become more extensive than usual by temperature scaling, increasing the probability of being selected as the next token. When the correct answer logit does not have a high value, the possibility of being selected as the next token is crucially dropped. For a low-temperature scale, once the model starts generation with an incorrect token, it is more likely to continue generating incorrect responses. As mentioned in CD, in the case of high confidence in high logit sampling methods in a generation, a wrong token selection can significantly impact the quality of future responses. In this situation, using \alg can increase the likelihood that a low correct answer token will be selected as the correct answer through CD using contrast VA. As a result, it shows high robustness against the temperature sampling scale and increases the likelihood of providing an appropriate response.

\section{Selection Strategy}
\label{app:selection}
\myparagraph{Removing noisy augmentations via acceptence threshold.}
Using the distance $D$, we expect to select a VA that shows high-performance improvement when used on CD. However, there may exist cases where some VAs cannot be appropriate contrastive augmentation for a specific task overall. In this case, these VAs contribute less to performance improvement than other VAs on average and can sometimes become noise that prevents other VAs from being used as contrast. We use the Acceptance Threshold, a simple baseline that eliminates the noise VAs. To discover the suitableness of VAs for the target task, in the sample sub-dataset, we utilize the LVLM's first token generation distance by \alg for each VA. Let $c_{i}$ be the number of times that VA$_{i}$ selected as contrast VA among a total of $M$ VAs. For the $N$ data samples and acceptance threshold $\tau$, candidate VAs with $c_{i} < \tau \frac{N}{M}$ are treated as unsuitable for this task and removed. Throughout the main experiments, we used the acceptance threshold of $\tau=0.5$.

\section{Experiment Details}

\subsection{Experiment computation resource.}
\label{app:exp_resources}

In this paper, all reported our experiment used LVLM models can run on a single 48 GB NVIDIA RTX A6000. In the process of applying \alg, our model requires inference as the number of VAs used in the first step only, and each subsequent generation step requires twice token generation stages.

%% file: tab/mme_modelsize.tex
\begin{table*}[t]
\caption{
MME performance by different model sizes. 
}
\label{table:mme_modelsize}
\small
\centering
\renewcommand{\arraystretch}{1.1}
\resizebox{\textwidth}{!}{
\setlength{\tabcolsep}{3pt}{
\begin{tabular}{l|l|c|c|c|c}
\thickhline
Method        & Aug            & LLaVA-1.5 7B & LLaVA-1.5 13B & InstructBLIP 7B & InstructBLIP 13B \\
\hline
Regular &  -          & 1272.22 & 1327.55 & 1155.26 & 1151.45 \\
\hline
VCD     & noise mask  & 1323.44 & 1354.34 & 1218.90 & 1208.44 \\
\hline
\multirow{6}{*}{\shortstack{Single}}  & color       & 1347.24 & 1414.53 & 1224.26 & 1237.71 \\
        & edgetexture & 1350.68 & 1424.20 & 1221.15 & 1220.63 \\
        & sharpness   & 1323.60 & 1361.74 & 1177.84 & 1164.32 \\
        & randcrop    & 1338.50 & 1384.65 & 1194.13 & 1205.55 \\
        & randerase   & 1310.89 & 1365.91 & 1195.27 & 1185.32 \\
        & flip        & 1344.75 & 1374.62 & 1222.34 & 1213.61 \\
\hline
 \multirow{2}{*}{\shortstack{VACoDe}} & all-in-one  & \textbf{1368.89} & \underline{1443.06} & \underline{1249.56} & \underline{1248.30} \\
        & selection   & \underline{1364.36} & \textbf{1443.14} & \textbf{1254.16} & \textbf{1256.09}  \\
\thickhline
\end{tabular}
}
}
\end{table*}

%% file: tab/mme_sampling.tex
 \begin{table*}[t]
\caption{
MME Performance by different sampling strategies.
}
\label{table:mme_sampling}
\small
\centering
\renewcommand{\arraystretch}{1.1}
\setlength{\tabcolsep}{3pt}{
\begin{tabular}{l|l|c|c|c|c}
\thickhline
\multirow{2}{*}{Method}        & \multirow{2}{*}{Aug}             & Top P & Top K & \multicolumn{2}{c}{Temperature}  \\
 & & $p=0.9$ & $k=50, T=0.7$ & $T=0.7$ & $T=1.5$ \\
\hline
Regular &  -          & $1352.87$ & $1399.33$ & $1403.99$ & $1169.71$ \\
\hline
VCD     & noise mask & $1370.47$ & $1425.60$ & $1429.52$ & $1316.95$ \\
\hline
\multirow{6}{*}{\shortstack{Single}}  & color      & $1405.90$ & $1443.27$ & $1445.19$ & $1349.20$ \\
        & edge       & $1434.14$ & $1433.86$ & $1420.64$ & $1364.72$ \\
        & sharp      & $1381.00$ & $1415.88$ & $1416.63$ & $1294.08$ \\
        & crop       & $1391.01$ & $1413.09$ & $1422.15$ & $1342.43$ \\
        & erase      & $1374.47$ & $1404.27$ & $1399.67$ & $1315.08$ \\
        & flip       & $1404.76$ & $1426.97$ & $1425.54$ & $1340.88$ \\
\hline
\multirow{2}{*}{\shortstack{VACoDe}}        & all & \textbf{1462.67} & \underline{1456.03} & \underline{1454.32} & \textbf{1389.03} \\
        & selection  & \underline{1462.58} & \textbf{1457.10} & \textbf{1458.73} & \underline{1377.47} \\
\thickhline
\end{tabular}
}
\end{table*}